\pdfoutput=1
\documentclass[11pt]{article}

\usepackage[]{acl}

\usepackage{times}
\usepackage{latexsym}

\usepackage{makecell}
\usepackage[T1]{fontenc}
\usepackage[utf8]{inputenc}

\usepackage{microtype}

\usepackage{lipsum} 
\usepackage{enumitem}
\usepackage{graphicx}
\usepackage{booktabs}
\usepackage{multirow}

\usepackage{CJK}

\usepackage{hyperref}

\usepackage{colortbl}
\definecolor{aliceblue}{RGB}{178, 217, 245}

\definecolor{babyblue}{RGB}{217, 239, 251}

\definecolor{babypink}{RGB}{251, 231, 230}
\definecolor{mygreen}{HTML}{3cb44b}

\newlength\myheight
\newlength\mydepth
\newcommand{\BenchName}{MoleculeQA}

\usepackage{tcolorbox}
\usepackage[linesnumbered,ruled,vlined]{algorithm2e}
\SetKwComment{Comment}{\color{green!50!black}\# }{}

\newcommand{\var}{\texttt}

\SetKwProg{Function}{def}{:}{}

\SetKwProg{For}{for}{:}{}
\SetKwProg{If}{if}{:}{}
\newcommand{\VarSty}[1]{\textnormal{\ttfamily\color{blue!90!black}#1}\unskip}

\newcommand\blfootnote[1]{%
	\begingroup
	\renewcommand\thefootnote{}\footnote{#1}%
	\addtocounter{footnote}{-1}%
	\endgroup
}

\title{\BenchName: A Dataset for Molecular Factual Question Answering}

\title{\BenchName: Unveiling the Factual Precision of 
Language Models in Molecular Comprehension}

\title{\BenchName: A Dataset to Evaluate Factual Accuracy in Molecular Comprehension}

\author{
Xingyu Lu$^{1,2, \dagger}$\quad He Cao$^{1,3 \dagger}$\quad Zijing Liu$^{1}$\quad Shengyuan Bai$^{1}$ \\ \textbf{Leqing Chen}$^{1}$\quad  \textbf{Yuan Yao}$^{3}$\quad \textbf{Hai-Tao Zheng }$^{2} \ddagger$\quad \textbf{Yu Li}$^{1}\ddagger$\\
$^{1}$International Digital Economy Academy (IDEA)~~~~~~\\$^{2}$Tsinghua Shenzhen International Graduate School, Tsinghua University~~~~~~\\ $^{3}$Hong Kong University of Science and Technology
}

\usepackage{amsmath,amsfonts,bm}

\def\eqref#1{equation~\ref{#1}}
\def\1{\bm{1}}

\DeclareMathAlphabet{\mathsfit}{\encodingdefault}{\sfdefault}{m}{sl}
\SetMathAlphabet{\mathsfit}{bold}{\encodingdefault}{\sfdefault}{bx}{n}

\usepackage[capitalize]{cleveref}

\begin{document}
\maketitle
\begin{abstract}
\blfootnote{$\dagger$ Equal contribution. Work is done during their internship at IDEA.}
\blfootnote{$\ddagger$ Corresponding author: Hai-Tao Zheng and Yu Li (E-mail: zheng.haitao@sz.tsinghua.edu.cn, liyu@idea.edu.cn)}
% \blfootnote{$\ddagger$ Corresponding authors.}
% \lipsum[1]
% In this paper, we introduce \BenchName{}, a novel question answering (QA) dataset, to rectify the absence  of factual evaluation in the molecular domain. The task of \BenchName{} is to answer questions about specific aspects of molecules (e.g., \textit{Which kind of antiviral activity does this molecule have/exhibit?}) by identifying the correct option from multiple choices. To construct \BenchName{}, we build a taxonomy to gather molecular facts into specific topics. With questions by topic, we combine powerful LLM and human evaluation to generate reliable answers. \BenchName{} has in total 62K QA pairs over 23K molecules, each QA pair includes a manually designed question, a positive option and three negative options, and corresponds to a molecular description statement with consistent semantics. \BenchName{} is not only the first benchmark for molecular factual bias evaluation but also the largest QA dataset in the molecular domain. A comprehensive evaluation based on \BenchName{}~\footnote{Our dataset and codes will be public after peer review.} for existing molecular LLMs exposes their deficiencies in specific areas and pinpoints several particularly crucial factors for molecular understanding.
Large language models are playing an increasingly significant role in molecular research, yet existing models often generate erroneous information, posing challenges to accurate molecular comprehension. Traditional evaluation metrics for generated content fail to assess a model's accuracy in molecular understanding. To rectify the absence of factual evaluation, we present \BenchName \footnote{https://github.com/IDEA-XL/MoleculeQA}, a novel question answering (QA) dataset which possesses 62K QA pairs over 23K molecules. Each QA pair, composed of a manual question, a positive option and three negative options, has consistent semantics with a molecular description from authoritative molecular corpus. \BenchName{} is not only the first benchmark for molecular factual bias evaluation but also the largest QA dataset for molecular research. A comprehensive evaluation on \BenchName{} for existing molecular LLMs exposes their deficiencies in specific areas and pinpoints several particularly crucial factors for molecular understanding.

\end{abstract}
% Abstract from ChatGPT
% In molecular research, language models (LMs) play a crucial role, but existing models often generate incorrect information, posing challenges in accurate comprehension. Current benchmarks, relying on n-gram metrics, lack the ability to assess the deeper issue of model understanding in the molecular domain, risking misinformation and limiting positive applications. To address this, we present \textbf{CUB}, a concise benchmark structured within a domain taxonomy that evaluates language models' comprehension of molecular knowledge, offering a nuanced assessment. Through testing on various models, we identify deficiencies, highlighting the need for improved benchmarks to enhance accuracy in the molecular domain.

% 1. 指出分子场景下大语言模型的发展现状和应用价值
% 2. 指出现存的模型生成内容中可能存在事实性错误，以及该领域对于模型的分子理解是否准确缺乏必要的评估。
% 3. 事实性错误可能导致的不良后果以及人工检测的高成本问题
% 4. 我们的数据集，构建流程，以及后续发现。

\section{Introduction}
\label{sec:Introduction}

Large Language Models (LLMs) have practical applications
%in scientific fields, including biomedical \cite{Med-Palm2, PubMedGPT}, meteorology \cite{ClimateBert}, and material science \cite{MatSciBERT}. One area where LMs have shown notable success is
in molecule research \cite{MolGPT, MolBert}, specifically in the use of cross-modal models \cite{MolCA, InstructMol}. These models bridge the gap between molecular structures and natural language \cite{Bran2023TransformersAL}, helping experts understand the properties, potential applications, and acquisition methods of specific compounds,
% They help predict reactions \cite{Schwaller_Laino_Gaudin_Bolgar_Hunter_Bekas_Lee_2019}, retrosynthesis \cite{Schwaller_Petraglia_Zullo_Nair_Haeuselmann_Pisoni_Bekas_Iuliano_Laino_2020}, and conditional de novo generation \cite{Text2Mol}, 
improving understanding of the complex molecular space and reducing experimental failure rates. 

\begin{table}[htb]
\small
\vskip -0.07in
\scalebox{0.90}{
\begin{tabular}{p{0.8cm}p{3.3cm}p{3.1cm}}
    \toprule
    \textbf{CID} & \textbf{Ground-Truth} & \textbf{Generated} \\
    \midrule
    \scriptsize{9810996} & \scriptsize{The molecule is a dipeptide composed of N-(3,3-dimethylbutyl)-\textcolor{black}{L-aspartic acid and methyl L-phenylalanate units joined by a peptide linkage.}} & \scriptsize{The molecule is a dipeptide obtained by formal condensation of the \textcolor{red}{alpha-carboxy group} of N-(3,3-dimethylbutyl)-L-phenylalanine \textcolor{red}{with ethanol}.} \\
    \midrule
    \scriptsize{10129879} & \scriptsize{The molecule is the stable isotope of potassium with relative atomic mass 38.963707.} & \scriptsize{The molecule is the stable isotope of \textcolor{red}{tellurium} with relative atomic mass \textcolor{red}{124.904425}.} \\
    \midrule
    \scriptsize{5281034} & \scriptsize{A synthetic androgen, it was mainly used for the treatment of anaemias until being replaced by treatments with fewer side effects.} & \scriptsize{It has a role as a \textcolor{red}{contraceptive drug, a progestin and a synthetic oral contraceptive}}. \\
    \midrule
    \scriptsize{15011611} & \scriptsize{It is found in Tripterygium wilfordii and Tripterygium hypoglaucum.} & \scriptsize{The molecule is an abietane diterpenoid \textcolor{red}{isolated from the stem bark of Fraxinus sieboldiana.}} \\
    \bottomrule
\end{tabular}
}
\vskip -0.1in
\caption{\footnotesize
The accuracy of the state-of-the-art molecule caption model in describing the given molecule is assessed using BioT5-base \cite{BioT5} inference results. The generated content is plausible and fluent, but comparison with the ground truth reveals several factual errors highlighted in \textcolor{red}{red}.
}
\vskip -0.1in
\label{tab:IntroAssess}
\end{table}

Although molecular LLMs are capable of generating task-relevant content, they often suffer from the problem of generating false statements.
In the widely studied molecule captioning task, users expect the model to generate a comprehensive and detailed description of a given compound. In this task, existing benchmarks \cite{MolT5, MolCA} typically employ n-gram metrics such as BLEU \cite{BLEU} and ROUGE \cite{ROUGE} to evaluate the performance of molecular LLMs. However, without examining the factual accuracy of these models, it is vague to justify how reliable they are. In~\autoref{tab:IntroAssess}, we provides several examples from the CheEBI-20 \cite{Text2Mol} test dataset to illustrate this issue. Despite the plausible and fluent appearance of the generated content, there are numerous unnoticed inaccurate statements, which remain difficult to detect under the current lexical-based benchmarking approach.

Counterfactual molecular generation content can lead to the following adverse consequences: 1) Misuse of deployed models can deceive and mislead ordinary users, reducing productivity. 2) Professionals may lower their expectations of deployed models when they recognize significant factual biases, thus hindering positive applications. To avoid these repercussions, quantifying the level of comprehension that models have of molecule knowledge is valuable. However, expertise and professional knowledge are required for human to detect hallucinations in generated molecular text, which is extremely difficult with high cost. 
% 3) Malicious actors can exploit these models to generate plausible false statements, intentionally deceiving users.
 
To alleviate this, we propose \textbf{\BenchName}, a comprehensive benchmark based on questions and answers covering various aspects such as property, source, structure, and application. \BenchName~endeavors to rectify the absence of reliable assessments of molecular knowledge comprehension of language models within the molecular domain.
% address two key issues: Firstly, the molecule-oriented domain lacks a comprehensive and reliable NLU benchmark to assess model understanding of molecule knowledge. Secondly, existing NLG benchmarks, such as molecule captioning, fail to detect model biases specific to the domain.

Construction of \BenchName~ involves two main stages. 1) \textbf{Domain Taxonomy Construction.} We utilize authoritative molecule description corpus as the source. Using a hybrid approach of rule-based and automated methods, we extract topics based on properties, sources, and other relevant aspects. After clustering and manual normalization, we gather the topics to build a hierarchical domain taxonomy that has broad coverage and strong expertise. 2) \textbf{Taxonomy-guided QA construction.} By converting each molecular description into several pairs of QA that align with the topics at different levels of taxonomy, we can create a QA benchmark that guarantees both granularity, breadth, and quality. \BenchName~ is not only the first factual evaluation benchmark in the molecular domain, but also the largest molecular QA dataset. Furthermore, we performed fine-tuning and accuracy tests on various molecular LLMs on \BenchName. Our experimental results indicate that existing methods remain at a discernible remove from achieving a precise comprehension of molecules, and undercover several vital factors for molecule modeling.
%\lyh[I think the process of constructing the dataset is too detailed. It is better to save space to introduce the advantages and characteristics of the dataset, as well as the insights obtained from the experiment.]c{}
Our contributions are summarized as follows:
\begin{itemize}[leftmargin=*]
    \setlength{\itemsep}{0pt}
    \item We have observed that the current language model in the molecule or chemistry domain exhibits factual bias in its descriptions of compounds, which cannot be adequately detected using existing metrics based on lexical similarity.
    \item To evaluate this bias, we have developed a domain taxonomy for molecule corpus and used it to create a high-quality comprehensive quality assurance benchmark called \BenchName.
    \item Using \BenchName, we have tested a series of models to assess their level of comprehension in the molecule domain. Based on our experimental outcomes, we identify specific deficiencies present in molecular models and summarize several critical factors for molecular understanding.
\end{itemize}

\section{Related Work}
\label{sec:related_work}

\subsection{Molecule Understanding LLMs}
Recent advancements in language models pre-trained with biomedical scientific corpora \cite{BioBERT,BioGPT,SciBERT} have shown considerable success in molecular research. Recently, cross-modal models have emerged \cite{Text2Mol, Text+ChemT5, MolFM, MolReGPT, molxpt}, aiming to bridge the gap between molecular language (bio-sequence or structure) and natural language. Evaluation tasks for these models include seq2seq generation-based tasks (e.g., molecule captioning and text-based de novo molecule generation) and contrastive-based tasks (e.g., cross-modal retrieval). The corresponding models can be classified as generative models (e.g., MolT5 \cite{MolT5}, BioT5 \cite{BioT5}) and contrastive models (e.g., MoMu \cite{MoMu}, MoleculeSTM \cite{MoleculeSTM}). 
% In addition, these models can also perform traditional single-modality tasks, such as molecular property prediction.

Seq2seq tasks assess the model's translation ability between modalities. Metrics evaluate the similarity between generated and ground truth content. For text-based de novo molecule generation, metrics include molecule fingerprint similarity (e.g. MACCS-FTS \cite{MACCS}, RDK-FTS, Morgan-FTS \cite{Morgan}), sequence-based metrics like BLEU \cite{BLEU} and validity. Molecule captioning tasks rely on n-gram precision (BLEU), recall (ROUGE \cite{ROUGE}), or both (METEOR \cite{METEOR}) to measure lexical similarity but lack chemical knowledge comparison and factual bias detection. Retrieval-type tasks align molecules with descriptions, but overlook fine-grained alignment and connections between text snippets and substructures. \BenchName{} introduces a hierarchical question-answering benchmarking framework, enabling a comprehensive evaluation of a model's molecular-related knowledge inferencing ability.

\subsection{Domain-Specific QA}
The Question Answering (QA) task serves as a quantitative measure for evaluating the reasoning and inference capabilities of intelligent systems. In the general domain, a large number of annotated QA samples have been constructed \cite{SQuAD, RACE, HotpotQA}. In addition, specific domains such as medical \cite{PubMedQA, MedQA, MedMCQA}, news \cite{CNNDailyMail, NewsQA}, and legal \cite{CaseHOLD, JEC-QA} have also developed standard QA datasets that are widely accepted and used by the community. QA data sets in specific domains can be classified into extraction-based \cite{BioRead, BIOASQ}, generation-based \cite{MEDIQA-AnS}, multichoice \cite{MedMCQA} and Yes / No formats \cite{PubMedQA}. QA pairs are constructed from various sources, including scientific articles \cite{PubMedQA}, examination questions \cite{MedQA, MedMCQA, MaScQA}, professional databases \cite{DrugChat}, and crowd-sourcing \cite{ChemistryQA, MMLU}. 

However, in the molecular domain, there is a scarcity of comprehensive, diverse, and high-quality QA datasets. Existing datasets like DrugChat \cite{DrugChat} have limitations in terms of molecule features and simplistic answers. 
BioMedGPT \cite{BioMedGPT} transforms molecule caption task datasets into QA samples, inheriting current evaluation issues like domain knowledge deficiency and excessive reliance on lexical similarity. Conversely, \BenchName{} constructs a domain taxonomy and derives QA pairs from descriptive texts, ensuring comprehensive, diverse, high-quality, and credible coverage. This makes it an effective benchmark for evaluating a model's understanding of molecular-oriented knowledge.

\section{Method}
\label{sec:Method}
% In this chapter, we will describe in detail how to construct MoleculeQA: The construction of MoleculeQA includes two processes: (1) Taxonomy construction, where a molecular description oriented taxonomy can organize specialized data, guiding question design and fine-grained evaluation. (2) Dataset construction, for each topic in the taxonomy, we obtain the final QA pairs through three steps: question design, answer collection, and quality evaluation. We will introduce them separately in sections 3.1 and 3.2

\subsection{Exposure of Factual Bias}
In this subsection, we analyze the presence of factual bias in the generated content of the molecule caption models.

\noindent
\textbf{Setup.} To evaluate the reliability of compound descriptions generated by these models, we categorized them into different aspects: \textit{Structure}, \textit{Property}, \textit{Application}, and \textit{Source}. The aspects were derived from descriptions in PubChem \cite{PubChem}, the largest molecule caption dataset currently available. PubChem includes specific sources for each molecule's description, such as Lotus \cite{Lotus} for source information, DrugBank \cite{DrugBank} for application details, CAMEO Chemicals \cite{cameochemicals} for property descriptions, and multiple data repositories for structure information. The definitions of these main aspects are summarized in the \cref{tab:Aspects_define} below.

\begin{table}[ht]
\small
\vskip -0.07in
\scalebox{0.9}{
\begin{tabular}{lp{5.5cm}}
    \toprule
    \textbf{Aspect } & \textbf{Definition}\\
    \midrule 
    \raisebox{-.5\height}{\includegraphics[width=0.06\linewidth]{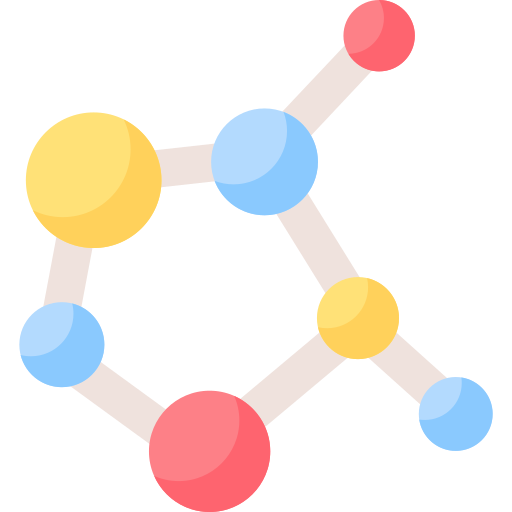}}\; Structure & {\small Details about architecture, composition, and interaction of atoms within a molecule.} \\
    \raisebox{-.5\height}{\includegraphics[width=0.06\linewidth]{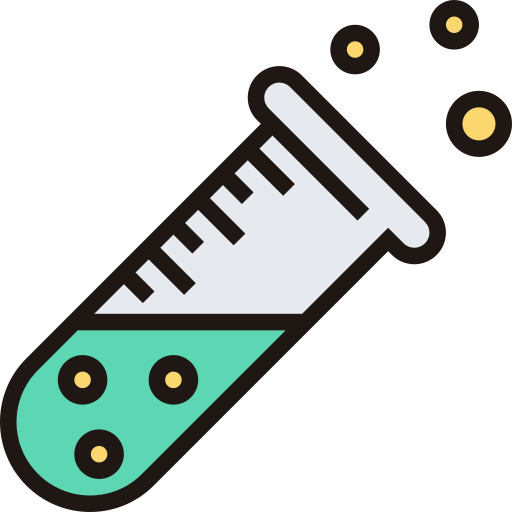}}\; Property & \small{Physical, biological or chemical property in various environments or reactions.} \\
    \raisebox{-.5\height}{\includegraphics[width=0.06\linewidth]{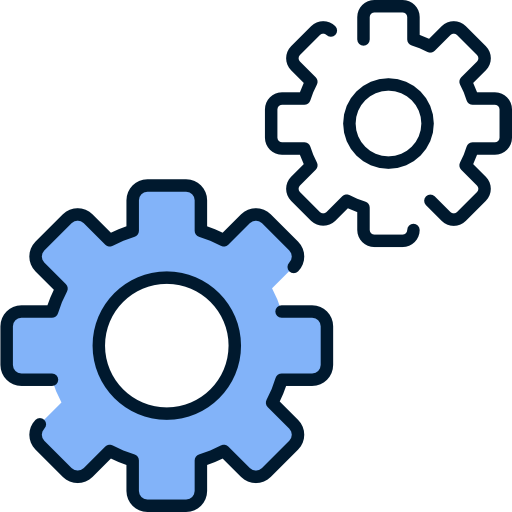}}\; Application & \small{The utilization of a molecular compound in various applications and scenarios.}\\
    \raisebox{-.5\height}{\includegraphics[width=0.06\linewidth]{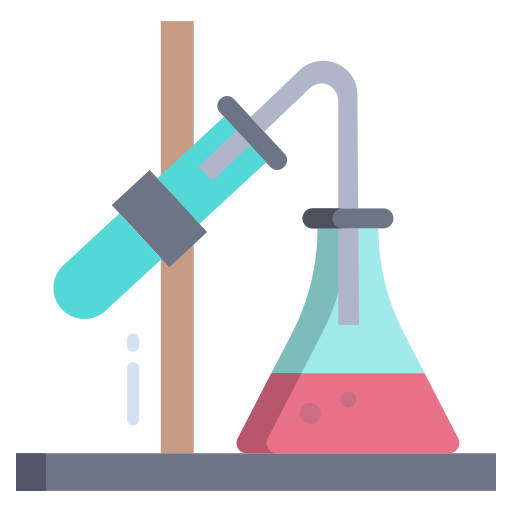}}\; Source & \small{The natural or synthetic origin, as well as the production context related to a molecule.} \\
    \bottomrule
\end{tabular}
}
\vskip -0.1in
\caption{\footnotesize
\textbf{Evaluation Aspects} of description about molecules.
}
\vskip -0.1in
\label{tab:Aspects_define}
\end{table}

We randomly sampled 100 molecule\&caption samples from the ChEBI-20 test set and used MolT5, MoMu, and BioT5 models to generate descriptions for each molecule. Both ground truth and generated content were manually classified based on four aspects. We evaluated the models' descriptions in each aspect against the ground truth, with two trained domain experts judging them as \textbf{{correct}} (if the generated content matched the ground truth), \textbf{miss} (if the ground truth had a corresponding aspect description but it was completely missing in the generated content), or \textbf{error} (if there was a clear factual inconsistency with the ground truth). 

\begin{figure}[!ht]
\vskip -0.1in
    \centering
    \includegraphics[width=0.95\linewidth]{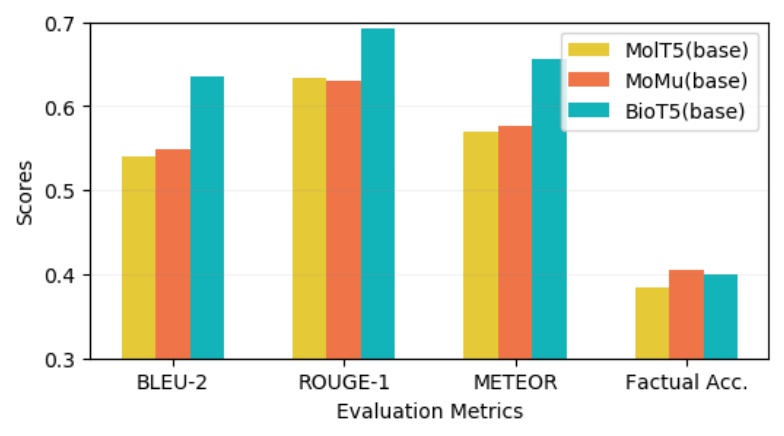}
    \vskip -0.12in
    \caption{\footnotesize{The performance of three representative models on the traditional metrics for the molecule caption task (e.g. BLEU etc.) and the factual accuracy metric we defined.}}
    \vskip -0.1in
    \label{fig:preExp}
\end{figure}

\begin{table}[ht]
\vskip -0.1in
\small
\scalebox{0.9}{
\begin{tabular}{lcccc}
    \toprule
     % & \multicolumn{4}{c}{error / miss / correct} \\
    \textbf{Model} & Structure & Property & Application & Source \\
    \midrule 
    MolT5-base &\textcolor{red}{63}/\textcolor{orange}{0}/\textcolor{mygreen}{34} &\textcolor{red}{1}/\textcolor{orange}{4}/\textcolor{mygreen}{3} &\textcolor{red}{7}/\textcolor{orange}{15}/\textcolor{mygreen}{8} &\textcolor{red}{20}/\textcolor{orange}{10}/\textcolor{mygreen}{30} \\
    % \midrule
    MoMu-base &\textcolor{red}{63}/\textcolor{orange}{0}/\textcolor{mygreen}{34} &\textcolor{red}{1}/\textcolor{orange}{4}/\textcolor{mygreen}{3} &\textcolor{red}{5}/\textcolor{orange}{16}/\textcolor{mygreen}{9} &\textcolor{red}{19}/\textcolor{orange}{\ 8}/\textcolor{mygreen}{33} \\
    % \midrule
    BioT5-base &\textcolor{red}{62}/\textcolor{orange}{0}/\textcolor{mygreen}{35} &\textcolor{red}{2}/\textcolor{orange}{3}/\textcolor{mygreen}{3} &\textcolor{red}{9}/\textcolor{orange}{12}/\textcolor{mygreen}{9} &\textcolor{red}{16}/\textcolor{orange}{13}/\textcolor{mygreen}{31} \\ 
    \bottomrule
\end{tabular}
}
\vskip -0.1in
\caption{\footnotesize \textbf{Human Assessment of Model Generated Molecular Descriptions} based on 4 aspects, with the counts presented according to \textcolor{red}{error} / \textcolor{orange}{miss} / \textcolor{mygreen}{correct}.
}
\vskip -0.1in
\label{tab:PreExp}
\end{table}

\begin{figure*}[!htbp]
    \centering
    \includegraphics[width=0.95\linewidth]{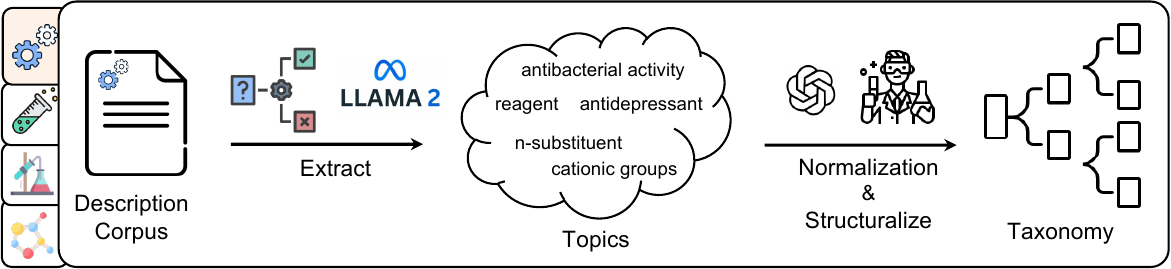}
    \vskip -0.1in
    \caption{\footnotesize{The process of constructing a molecular domain taxonomy. The procedures involve the selection of the information source, extraction of topics, normalization and structuralization of topics, and hierarchical clustering by domain experts.}}
    \label{fig:taxonomy}
    \vskip -0.1in
\end{figure*}

\begin{figure*}[!htbp]
    \centering
    \includegraphics[width=0.97\linewidth]{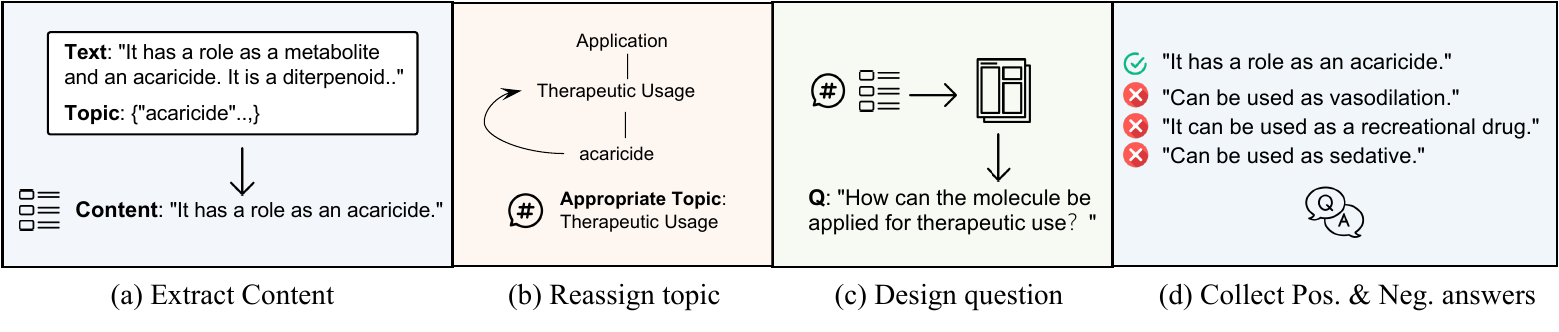}
    \vskip -0.15in
    \caption{\footnotesize{The process of constructing a molecular domain taxonomy. The procedures involve the selection of the information source, extraction of topics, normalization and structuralization of topics, and hierarchical clustering by domain experts.}}
    \vskip -0.1in
    \label{fig:QA_construction}
\end{figure*}

\noindent
\textbf{Results.} In Figure \ref{fig:preExp}, we assess the content generated by the model using traditional lexical-based metrics (BLEU, ROUGE, METEOR), as well as measure their factual accuracy on the selected subset. We define factual accuracy as the ratio of correct predictions to the total number of slots, serving as an average metric to evaluate the reliability of the generated content. Despite the progress in training methodologies, models have exhibited incremental improvements in lexical similarity metrics (such as a 17.6\% increase in BLEU-2). Nevertheless, there has been no discernible improvement in the dependability of the generated content, with factual accuracy persisting at 0.4. In our detailed factual performance analysis (Table \ref{tab:PreExp}), we observed models often omit application-related details and relevant properties. Describing molecular structures showed significant discrepancies of more than 63\% compared to ground truth. This challenges the credibility of expert model-generated content, which warrants further scrutiny.

\subsection{Domain Taxonomy Construction}
\label{TaxCon}

Taxonomy frameworks organize concepts or entities within a domain hierarchically, aiding in the organization of domain-specific queries \cite{Liu2012AutomaticTC} and ensuring the quality of comprehension domain knowledge and constructing question-answering pairs. We adhere to established procedures for the construction of domain taxonomies, as illustrated in Figure \ref{fig:taxonomy}.

\noindent
\textbf{Information Source.} Considering the data quality, we choose the most widely used ChEBI-20 dataset as our molecular description corpus. To mitigate the class imbalance issue in ChEBI-20, primarily dominated by structural information, we include additional sources like T3DB \cite{T3DB}, FDA Pharm Classes, and DrugBank. We employ a pre-trained text classifier to perform an initial coarse-grained division of the corpus based on the four aspects we defined above, which serve as the first-level nodes in our taxonomy.

\noindent
\textbf{Topics Extraction.} We further employ a hybrid approach combining rule-based and few-shot prompting methods to extract topics and their corresponding original text from the corpus, formatting the $(topic, text)$ pairs. Subsequently, to mitigate lexical noise and uncontrolled granularity within the 1K topics collected, we utilize GPT-4 \cite{gpt4v} with a few-shot prompt-based approach to accomplish an initial semantic aggregation.

\noindent
\textbf{Topics Normalization \& Structuralization.} Next, domain experts intervene to perform rule-based topic merging and concept splitting manually. Subsequently, the remaining 587 topics are hierarchically clustered by human experts, resulting in a three-level molecular domain taxonomy. An overview of this taxonomy can be found in the Appendix. The leaf nodes represent specific molecule characteristics and are the narrowest topics/concepts, while non-leaf nodes represent broader concepts.

\subsection{\BenchName{} Construction}
Based on the taxonomy in \ref{TaxCon}, we develop a 4-step procedure to extract questions and answers from molecular descriptions to construct \BenchName{}. The whole workflow is displayed in Fig \ref{fig:QA_construction}.

\begin{table*}[!ht]
\centering
\small
\scalebox{0.75}{
\begin{tabular}{p{1.3cm}p{5.6cm}p{3.2cm}p{4.6cm}p{4.2cm}}
    \toprule
    \textbf{Taxonomy} & \textbf{Reference Description} & \textbf{Extracted Question} & \textbf{Positive Answer} & \textbf{Negative Answer} \\
    \midrule
    Property$\rightarrow$ Antiviral activity & It has been shown to \textcolor{blue}{\textit{exhibit inhibitory effects on the viral neuraminidases from two influenza viral strains, H1N1 and H9N2}}. & Which kind of \textbf{antiviral activity} does this molecule have/exhibit? & It exhibits inhibitory effects on the viral neuraminidases from two influenza viral strains, H1N1 and H9N2. & \textcolor{red}{It is used for the treatment of cytomegalovirus (CMV) retinitis in AIDS patients.}\\
    \midrule
    Structure$\rightarrow$ Backbone & The molecule is a heparan sulfate composed of a backbone of \textcolor{blue}{\textit{repeating beta-D-glucuronosyl-(1->4)- N-sulfonyl-alpha-D-glucosamine units joined by (1->4)-linkages}}. & Which kind of \textbf{backbone} does this molecule have? & It has a backbone of repeating beta-D-glucuronosyl-(1->4)-N-sulfonyl-alpha-D-glucosamine units joined by (1->4)-linkages & It has a backbone of repeating \textcolor{red}{alpha-L-iduronosyl}-(1->4)-N-sulfonyl-alpha-D-glucosamine units joined by (1->4)-linkages. \\
    \bottomrule
\end{tabular}
}
\vskip -0.1in
\caption{\footnotesize Examples of automatically generated QA instances. \textcolor{blue}{\textit{blue}} stands for reference locations, \textcolor{red}{red} for factual errors. }
\vskip -0.1in
\label{tab:qa_example}
\end{table*}

\noindent
\textbf{Content Extraction \& Reassign Topic.}
With $(topic, text)$ pairs annotated in \ref{TaxCon}, a reasonable notion is to query molecules by topic, but content related to a specific topic can be over-brief to be queried. For example, for molecule \texttt{\small{CID:5479113}}, content of topic \texttt{\small{acaricide}} is \texttt{\small{It has a role as an acaricide}}. Without enough information, it is difficult to justify which species of mites this molecule is effective. However, it can be queried from a coarser granularity like \texttt{\small{Therapeutic Usage}}, the parent topic of \texttt{\small{acaricide}}. 

To select a suitable topic for querying, we first use an agent to extract \text{content} related to the  \text{topic} from \text{text}. A rule-based program is employed to verify the content and, in cases where specific details about a given topic are unavailable, we replace the topic with its parent topic until the level of granularity is appropriate for querying purposes. 

\noindent
\textbf{Question Design.} We invite two annotators to design questions for topics based on the extracted contents. For example, contents for topic \texttt{\small{inhibitor}} include \texttt{\small{It is a protein synthesis inhibitor}} and \texttt{\small{It is a mitotic inhibitor}}, annotators may design \texttt{\small{Which kind of inhibitor is this molecule?}}. For each topic, annotators discuss choosing the better design as its final question and make sure each question can be answered using the molecular descriptions.

\noindent
\textbf{Pos. Options Collection.}
% Explain questions for the same topic is consistent, the basis of negative sampling
% Generate the right answer
For the positive option, since formal extracted contents may be rigid and can't be directly used as answers, we leverage the in-context learning capability of ChatGPT \cite{Chatgpt} to generate appropriate positive options via few-shot prompting.

\noindent
\textbf{Neg. Options Collection.}
For the same question, we take positive options from other molecules as negative candidates for each molecule. To eliminate illegal negatives, we merge synonymous options and remove overlapping options. Then we adopt BioT5 \cite{BioT5} to encode all candidates and choose candidates with similar semantics to the positive option as negatives. Several generated QA instances are shown in Table \ref{tab:qa_example}.

\noindent
\textbf{Data Split.}
We split molecules in \BenchName{} into train/dev/test sets by scaffolds to divide molecules with similar structures into the same sets as suggested in \cite{Hu2019StrategiesFP}, making the QA task more challenging yet realistic.

\noindent
\textbf{Quality Control.}
To provide reliable factual evaluation, LLM and human efforts are combined to ensure \BenchName's quality. We convert each QA instance into natural language using templates and assess its logical and semantic consistency with the original description using ChatGPT. This process is repeated 3 times to minimize variations. With taxonomy guidance, the number of disqualified samples is minimal and can be manually resolved.

\noindent
\textbf{Human Evaluation.}
We assign one annotator~\footnote{All annotators are doctoral students engaged in molecule research, with at least six months of professional experience.} to evaluate the reliability of the test split and receive error rate lower than 1\%.
Finally, we randomly sample 100 cases and assign two annotators to evaluate the quality of QA samples. The annotators assess the \textbf{Consistency} between the question and the correct option with the reference caption text, as well as \textbf{Discrimination} between the positive and negative options. Human evaluation results can be found in Table \ref{tab:qa_quality}. The high consistency and discrimination metrics, along with a satisfactory level of agreement (Cohen kappa) among annotators, validate the quality and reliability of our \BenchName{}.

\begin{table}[!ht]
\centering
\small
\vskip -0.1in
\scalebox{0.85}{
\begin{tabular}{lccc}
    \toprule
    \textbf{Metric} & \textbf{Annotator 1} & \textbf{Annotator 2} & 
    \textbf{Agreement} ($\kappa$) \\
    \midrule
    \textbf{Consistency} & 99.0 & 99.0 & 1.0 \\
    \textbf{Discrimination} & 97.0 & 96.0 & 0.85 \\
    \bottomrule
\end{tabular}
}
\vskip -0.1in
\caption{\footnotesize Evaluation for the generated QAs quality.}
\label{tab:qa_quality}
\vskip -0.1in
\end{table}

\subsection{Data Analysis}
\noindent
\textbf{Data Statistics.} 
In \cref{tab:comparison}, we present the number of QA samples and the coverage of topics in \BenchName{} in comparison to several popular bio-molecular and chemistry-related benchmarks \cite{ChemistryQA, MMMU, MMLU, ScienceQA}. We observe that \BenchName{} is both the first benchmark focused on evaluating molecular factual knowledge and the largest scale QA dataset in the molecular field. 
\begin{table}[!ht]
\centering
\small
\vskip -0.07in
\scalebox{0.83}{
\begin{tabular}{lcc}
    \toprule
    \textbf{Benchmarks} & \textbf{\# QA } &\textbf{Sophistication} \\
    \midrule
    MMLU(Chem)  &534 & College, High school, Medicine\\
    MMMU(Chem) & 638 & Inorganic, Organic, Physical\\
    ScienceQA & 867& Solution, Reaction, Molecule \\
    ChemistryQA & 4,500 & Reaction, Molecule, Physics \\
    \midrule
    MoleculeQA & 61,574 &\begin{tabular}[c]{@{}c@{}}Structure, Source, Property, Application\end{tabular}  \\
    \midrule
\end{tabular}
}
\vskip -0.1in
\caption{\footnotesize Number of samples and topics coverage compared to popular related benchmarks.}
\vskip -0.1in
\label{tab:comparison}
\end{table}

The train, development, and test split consists of 49,993, 5,795 \& 5,786 QA samples. The general statistics of the dataset are summarized in Table \ref{tab:statistics}. 

\begin{table}[ht]
\centering
\small
\label{tab:statistic}
\vskip -0.05in
\scalebox{0.7}{
\begin{tabular}{lccccc}
    \toprule
    \textbf{Aspects} & {\textbf{Structure}} & \textbf{Property} & \textbf{Application} & \textbf{Source} & \textbf{Total} \\
    \midrule
    \# Train & 32,176 & 4,838 & 1,917  & 11,062  & 49,993 \\
    \# Dev & 3,314 & 698 & 558 & 1,225 & 5,795\\
    \# Test & 3,113 & 731 & 599 & 1,343 & 5,786 \\
    Avg. Q Tokens & 7.96 & 9.02 & 7.90 & 7.00 & 7.74 \\
    Avg. A Tokens & 9.50 & 10.98 & 11.93 & 7.96 & 9.42\\
    \midrule
\end{tabular}
}
\vskip -0.1in
\caption{\footnotesize
\BenchName{} dataset statistics, where Q and A represent the Question and Answer respectively.
}
\vskip -0.1in
\label{tab:statistics}
\end{table}

\begin{figure}[!ht]
\vskip -0.1in
    \centering
    \includegraphics[width=0.9\linewidth]{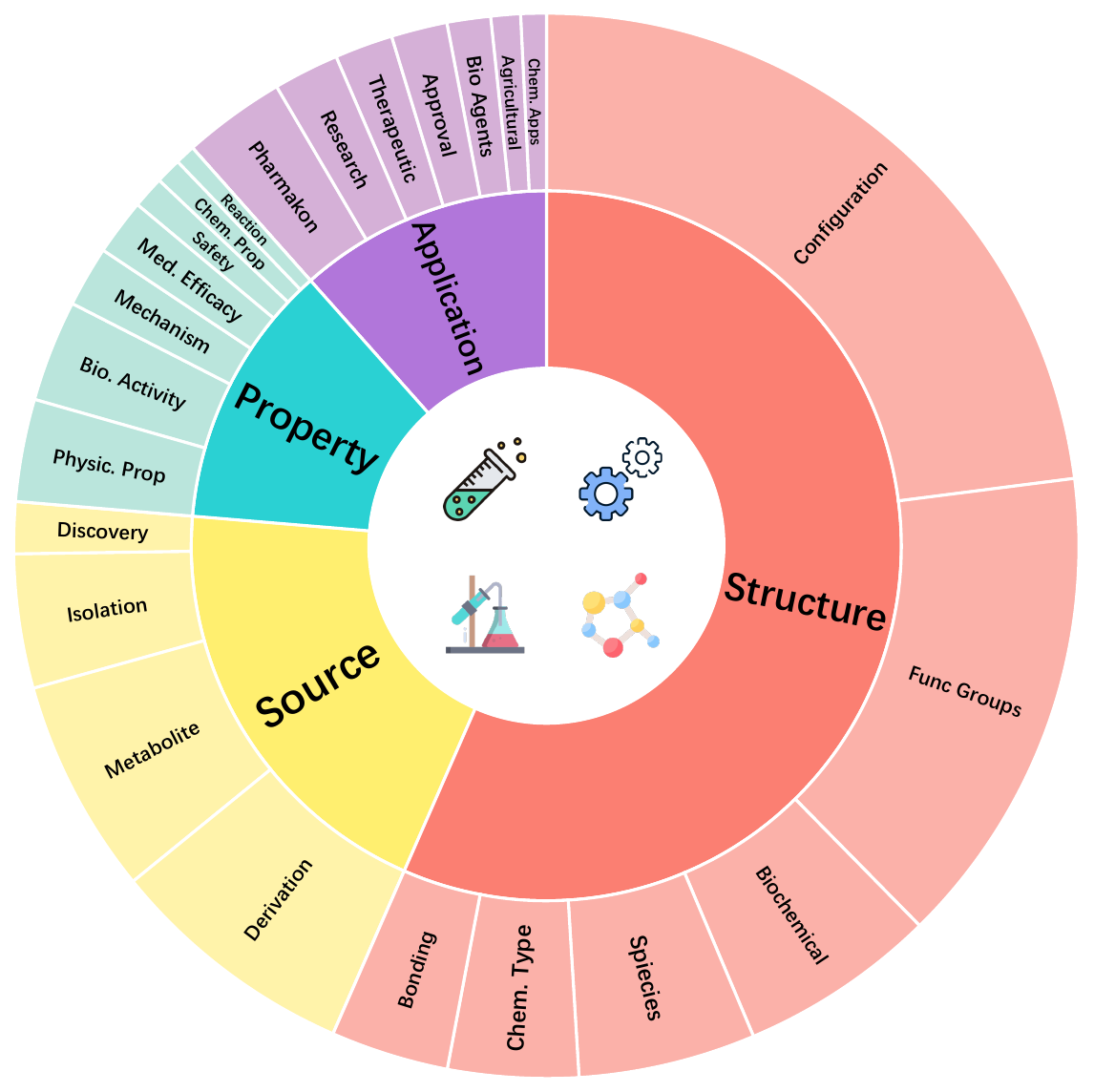}
    \vskip -0.1in
    \caption{\footnotesize{An overview of \BenchName{} topics distribution. Four coarse-grained  aspects occupy the inner circle, and 
in the outer circle we list finer-grained non-leaf topics.}}
    \vskip -0.1in
    \label{fig:qa_stats}
    \vskip -0.1in
\end{figure}

\noindent
\textbf{Data Distribution.} 
Fig \ref{fig:qa_stats} provides the visualized distribution of \BenchName{}. All topics in our taxonomy are queried in \BenchName{} for a comprehensive, fine-grained factual evaluation. Inherited from ChEBI-20, QA pairs in the \textit{Structure} aspect account for approximately two-thirds of the whole \BenchName{}. While topics within each aspect have relatively balanced sample numbers.

%TODO: paste new figure

% \begin{table}[ht]
% \centering
% \small
% \scalebox{0.8}{
% \begin{tabular}{lccccc}
%     \toprule
%     \textbf{Aspects} & \textbf{\#Train} & \textbf{\#Dev} & \textbf{\#Test} & \textbf{Avg. Q Tokens} & \textbf{Avg. A Tokens}\\
%     \midrule
%     Structure & 32,176 & 3,314 & 3,113 & 7.96 & 9.50 \\
%     Property & 4,838 & 698 & 731 & 9.02 & 10.98\\
%     Application & 1,917 & 558 & 599 & 7.90 & 11.93\\
%     Source & 1,1062 & 1,225 & 1,343 & 7.00 & 7.96 \\
%     \midrule
%     Total & 49,993 & 5,795 & 5,786 & 7.74 & 9.42\\
%     \midrule
% \end{tabular}
% }
% \vskip -0.1in
% \caption{\footnotesize
% \BenchName{} dataset statistics, where Q and A represent the Question and Answer respectively.
% }
% \vskip -0.1in
% \label{tab:statistics}
% \end{table}

\section{Experiment}
\label{sec:Experiment}

\begin{table*}[!ht]
    \vskip -0.25in
    \small
    \centering
    \setlength{\tabcolsep}{3mm}{
    \scalebox{0.85}{
    \begin{tabular}{lcccccccc}
        \toprule
        \multirow{1}{*}{\text{\textbf{Model}}} & 
        \multicolumn{1}{c}{\text{\textbf{\makecell[c]{\# Trainable\\ Params}}}} &
        \multicolumn{1}{c}{\text{\textbf{Implementation}}} &
        \multicolumn{1}{c}{\text{\textbf{Structure}}} &
        \multicolumn{1}{c}{\text{\textbf{Source}}} & 
        \multicolumn{1}{c}{\text{\textbf{Property}}} &
        \multicolumn{1}{c}{\text{\textbf{Application}}} &
        \multicolumn{1}{c}{\text{\textbf{Total}}} \\
        \midrule
        Random & -- & -- & 24.41 & 22.30 & 23.04 & 24.57  & 24.03  \\
        \midrule
        \rowcolor[RGB]{234, 238, 234}
        \multicolumn{8}{l}{\textit{Molecular LLM}} 
         \\
        \midrule
        \hspace{0.1cm}MolT5-small & 80M & full ft & 49.59 & 64.18 & 46.51 & 40.90 & 51.69  \\
        \hspace{0.1cm}MolT5-base & 250M & full ft  & 58.01 & 65.85 & 45.14 & 42.24 & 55.39  \\
        \hspace{0.1cm}MoMu-small & 82M & full ft & 52.71 & 63.44 & 44.87 & 40.57 & 52.96  \\
         \hspace{0.1cm}MoMu-base & 252M & full ft & 61.58 & 65.30 & 43.78 & 43.07 & 57.43  \\
         \hspace{0.1cm}BioT5-base & 252M & full ft  & 65.98 & 69.24 & \textbf{49.11} & 40.73 & 62.03  \\
          \hspace{0.1cm}MolCA-125M & 100M & LoRA ft & 65.54 & 67.34 & 45.77 & 40.33 & 60.30  \\
         \hspace{0.1cm}MolCA-1.3B & 110M & LoRA ft & \textbf{71.12} & \textbf{70.98} & 47.81 & \textbf{43.17} & \textbf{64.79}  \\
         \hspace{0.1cm}BioMedGPT-LM-7B & 40M & LoRA ft & 54.19 & 60.01 & 38.85 & 40.90 & 52.23 \\
        \midrule
        \rowcolor[RGB]{234, 238, 234}
        \multicolumn{8}{l}{\textit{General LLM}} 
         \\
        \midrule
        \hspace{0.1cm}T5-small & 60M & full ft & 55.51 & 64.41 & 45.42 & 38.56 & 54.55  \\
        \hspace{0.1cm}T5-base & 220M & full ft & \textbf{60.42} & \textbf{66.42} & 45.83 & \textbf{43.74} & \textbf{58.24}  \\
        \hspace{0.1cm}OPT-125M & 125M & full ft & 38.58 & 55.92 & 41.04 & 28.73 & 42.93  \\
        \hspace{0.1cm}OPT-350M & 331M & full ft & 44.39 & 60.83 & \textbf{46.24} & 40.57 & 48.05  \\
        % \hspace{0.1cm}Galactica-1.3B & 4.7M & LoRA ft & 29.39 & 37.75 & 27.22 & 30.22 & 31.14 \\
        \hspace{0.1cm}GALACTICA-6.7B  & 12.5M & LoRA ft & 32.35 & 41.92 & 31.05 & 28.21 & 33.96  \\
        % \hspace{0.1cm}BLOOM-1.7B & 11.0M & LoRA ft & 29.20 & 36.93 & 30.10 & 30.22 & 31.21 \\
        \hspace{0.1cm}BLOOM-7.1B  & 27.5M & LoRA ft & 35.01 & 47.51 & 31.46 & 33.56 & 37.31  \\
        % \hspace{0.1cm}Pythia-1.4B & 11.0M & LoRA ft & 41.25 & 54.58 & 40.08 & 40.07 & 44.07 \\
        \hspace{0.1cm}Pythia-6.9B  & 29.4M & LoRA ft & 42.79  & 58.90 & 38.58 & 39.07  & 45.61   \\
         \hspace{0.1cm}Mol-Instruction-7B & 40M & LoRA ft & 37.46 & 47.36 & 32.69 & 29.88 & 38.37 \\
        \hspace{0.1cm}Llama-2-7B-chat & 40M & LoRA ft & 28.75 & 39.84 & 31.33 & 27.71 & 31.54 \\
        \hspace{0.1cm}Llama-2-13B-chat & 63M & LoRA ft & 34.37 & 43.86 & 31.05 & 29.72 & 35.67 \\
        \hspace{0.1cm}Vicuna-v1.5-7B & 40M & LoRA ft & 34.89 & 44.15 & 34.20 & 31.55 & 36.61 \\
        \hspace{0.1cm}Vicuna-v1.5-13B & 63M & LoRA ft & 37.01 & 43.19 & 30.64 & 31.55 & 37.07 \\
        \midrule
        \rowcolor[RGB]{234, 238, 234}
        \multicolumn{8}{l}{\textit{Large-scale Universal Models}} 
         \\
        \hspace{0.1cm}Mixtral-8$\times$7B-Instruct-v0.1 & -- & 10-shot & 23.32 & 31.87 & 32.89 & 29.96 & 27.79 \\
        \hspace{0.1cm}GPT-3.5-1106-turbo & -- & 10-shot  & 25.60 & 37.60 & 28.04 & 32.22 & 29.29 \\
        \hspace{0.1cm}GPT-4-1106-preview & -- & 10-shot  & \textbf{60.94} & \textbf{50.19} & \textbf{35.57} & \textbf{43.91} & \textbf{53.47} \\ 
        \bottomrule
    \end{tabular}}
    }
    \vskip -0.15cm
    \caption{\footnotesize{We report the accuracy (\%) results on \BenchName{} test set under different aspects (\textbf{Best} for model-wise).}}
    \label{tab:main}
\end{table*}

\subsection{Baseline Models}
The main purpose of baseline experiments is to investigate current models' performance in answering multiple-choice questions related to molecular knowledge. We categorize models based on whether their base LLMs are adequately trained on a large-scale biomolecular corpus as follows: 

\noindent
\textbf{Molecular LLM}, represented by MolT5 \cite{MolT5}, MoMu \cite{MoMu}, BioT5 \cite{BioT5}, MolCA \cite{MolCA} and BioMedGPT-LM-7B \cite{BioMedGPT}. 
These models undergo multiple training rounds during the pretraining or incremental training stages, utilizing extensive molecular modality data (e.g. SMILES or SELFIES strings), biomedical-related scientific papers, and molecule-description pairs.

\noindent
\textbf{General LLM}, represented by T5 \cite{T5}, OPT \cite{OPT}, GALACTICA \cite{Galactica}, BLOOM \cite{BLOOM}, Pythia \cite{Pythia},  LLama-2 \cite{Llama2}, along with its instruction fine-tuned derivatives, such as Vicuna \cite{vicuna} and Mol-Instruction-7B \cite{Mol-Instructions}. 
% which are fine-tuned on user-shared ChatGPT conversations, and Mol-Instruction-7B \cite{Mol-Instructions}, which undergoes fine-tuning based on molecule-oriented instructions.

\noindent
\textbf{Large-scale Universal Models}. We evaluate the large-scale, state-of-the-art LLMs in few-shot settings, including open-access models such as Mixtral 8$\times$7B \cite{Mixtral8x7B}, and OpenAI's GPT family, specifically GPT-3.5 \cite{Chatgpt} and GPT-4 \cite{gpt4v} accessed via API~\footnote{https://api.openai.com/v1/chat/completions}.

\subsection{Evaluation Setups}
% ref: 
% 1. Metric intro; 在四个aspect衡量
% 2. 介绍training methods: Full fine-tuning; PEFT LoRA; Few-shot 
Training approaches in our evaluation include:

\noindent
\textbf{Full Fine-tuning}: All parameters of the whole model are updated, including the base LLMs, structure encoders, and projectors for aligning molecular structure and natural language modality.

\noindent
\textbf{LoRA-based Fine-tuning}: The base LLMs are tuned by low-rank adaptation \cite{LoRA}, and structure encoders are also trainable.

\noindent
\textbf{Few-shot Setting}: We sample 10 QA examples from four aspects respectively to prompt LLMs with task definition and contextual information.

We select training approaches and hyper-parameters consistent with the original papers for respective models. Details about training configuration and few-shot prompting examples are provided in the Appendix. 
The main metric of \BenchName{} is the \textbf{accuracy}, which is defined as the ratio of correctly answered samples among all test samples. 
We present the corresponding accuracy for four aspects as well as the total accuracy in Table \ref{tab:main}.

% \noindent
% \textbf{T5, MolT5, BioT5:} MolT5 \cite{MolT5} is built on the T5 \cite{T5} architecture and is trained on both SMILES and general text corpus. BioT5 \cite{BioT5} incorporates SELFIES for robust molecular representations and employs a text-wrapping method to extract information from the surrounding context of biological entities.

\subsection{Main Results}
We summarize the benchmarking results in \cref{tab:main}:
\begin{itemize}[noitemsep,topsep=0pt,leftmargin=*]
\item \textbf{Comparison over four aspects.} Achieving the highest accuracy on \textit{Source} is generally more feasible for each model, whereas addressing \textit{Property} and \textit{Application} presents notable difficulties across all models, with none surpassing a 50\% accuracy rate. This phenomenon may be ascribed to the comparatively smaller sample sizes within these particular domains.

\item \textbf{Molecular LLMs v.s. General LLMs.} Molecular LLMs demonstrate better performance, with a minimum total accuracy over 51\%. By contrast, other than T5s, decoder-only General LLMs fail to achieve a total accuracy exceeding 50\%, whether fully fine-tuned or tuned with LoRA.

\item \textbf{T5 series comparison.} Among T5-based methods, T5 demonstrates superior performance compared to MolT5 (e.g., T5-base achieves total accuracy surpassing MolT5-base by 5.1\%) contradicting their performance on molecule caption tasks. BioT5 combines bio-molecular texts and databases for molecular pretraining, achieving higher accuracy than T5 (+ 6.5\%).

\item \textbf{Decoder-only LLMs comparison.} 
Among Llama-based models, BioMedGPT-7B achieves the best performance by incremental pre-training, while Mol-Instruction fine-tuned by instructions has slight improvement than Llama and Vicuna.
With the similar size of the base model (7B) and LoRA parameters, the performance ranking among different models is as follows: Pythia > BLOOM > GALACTICA > Llama, which may provide a reference for molecular base model selection. Increasing model size (e.g. 7B$\rightarrow$13B) also receives mild accuracy gain.

\item \textbf{Single v.s. Multiple modalities.}
Both MoMu and MolCA are models that jointly incorporate molecular 2D graph modality and textual information. They demonstrate improvements over their base models (MolT5 and GALACTICA respectively) that solely rely on a single 1D-text modality.
% Best performance?
\item \textbf{Large-scale Universal Models} The utilization of highly advanced models, such as GPT-4, has potential in the field of molecular research. In a 10-shot scenario, GPT-4 demonstrates accuracy comparable to certain specialized models. However, the performance of smaller models declined sharply, which may be attributed to the lack of their emergent abilities(\cite{emergent}).
% 涌现？
\end{itemize}

\begin{figure*}[!t]
\centering
\vskip -0.2in
\includegraphics[width=1.0\textwidth]{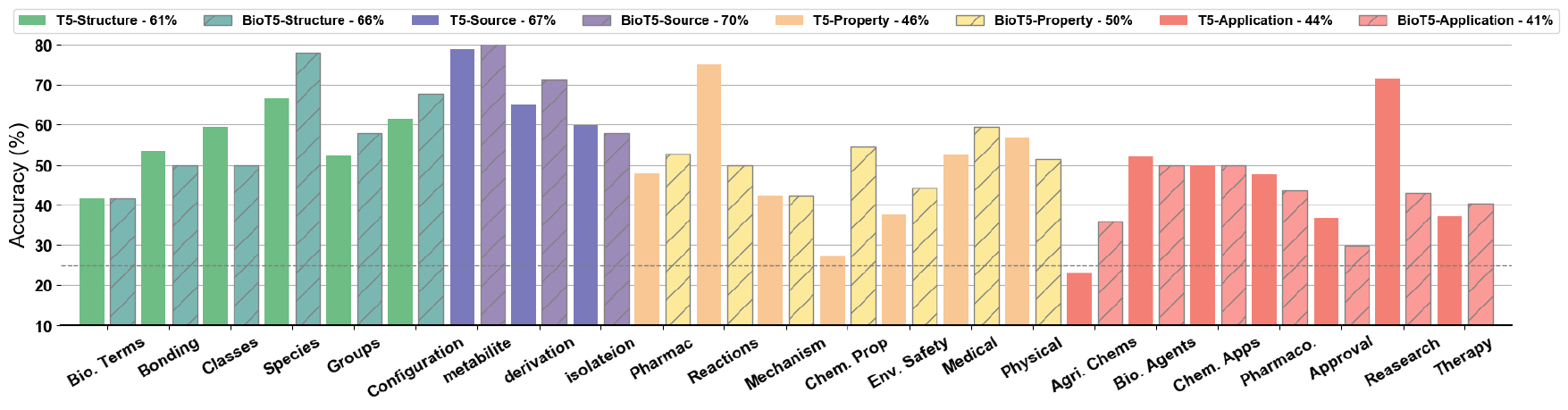}
\vskip -0.14in
\caption{\footnotesize
Accuracy of different finer topics under 4 coarse-grained aspects on the \BenchName{} test set. We select BioT5- and T5-base as representatives of Molecular LLM and General LLM, respectively, represented by solid and dashed bars.
}
\label{fig:topic_acc}
\vskip -0.2in
\end{figure*}

\section{Analysis}
% RQ-writing ref: https://arxiv.org/pdf/2311.04044.pdf
We propose the following research questions(RQs) for the molecular domain to guide our analysis:

\begin{itemize}[noitemsep,topsep=0pt,leftmargin=*]
\item\textbf{RQ1}: Are existing LLMs powerful enough for application in practical molecular scenarios?
\item\textbf{RQ2}: What factors are crucial for enhancing LLMs' ability for molecule comprehension? 
\item\textbf{RQ3}: How do LLMs adhere to the scaling law in molecular scenarios?
\end{itemize}

\subsection{In-depth Performance Analysis (RQ1)}
We draw a preliminary conclusion from \cref{tab:main} that existing LLMs' comprehension of molecules is far from satisfactory: When confronted with aspects of Property and Application, pivotal for real-world applications, evaluated models consistently fail to achieve commendable accuracy. To more thoroughly assess the methods' level of comprehension across various molecular aspects, we plot T5-base and BioT5's accuracy over each sub-category in our taxonomy in \cref{fig:topic_acc}. We find that in aspects of \textit{Source} and \textit{Structure}, two models exhibit consistent performance, with accuracy exceeding 40\% across all categories. But on sub-topics like \textit{Agricultural Chemical} and \textit{Approval status}, two models perform notably sub-optimal. Various accuracy on different topics can serve as a confidence coefficient for related model applications.

\subsection{Crucial Factor Attribution (RQ2)}
We summarize the following crucial factors for improving molecular comprehension ability:

\noindent
\textbf{Molecular Corpora.}
The two text-based variants derived from T5, MolT5 and BioT5, displayed divergent outcomes. MolT5 exhibited lower performance compared to T5, while BioT5 demonstrated improved performance. The observed performance divergence between MolT5 and BioT5 can be attributed to the differences in their training corpora, specifically in terms of scale and diversity.
Similarly, decoder-only models also exhibit this phenomenon: BioMedGPT (4.2M bio-molecular papers) > Mol-Instruction (1M molecular-oriented instruction samples) > Vicuna (70K general instruction samples) > Llama(General corpus). The above findings emphasize the importance of large, diverse, and high-quality datasets specific to the molecular domain for improving performance.

\noindent
\textbf{Modality Modeling Strategy.}
We investigate which modality modeling strategies can more effectively facilitate molecular modeling. (1) \textbf{Modality learning:} There is a significant performance gap between LoRA-based textual methods and methods employing multi-modal fusion or full fine-tuning, which underscores that a certain scale of trainable parameters is necessary to adequately model the textual or graph modalities of molecules. (2) \textbf{Multi-modal fusion:} Both MolCA and MoMu demonstrate that fusing molecular graphs into the semantic space of LLMs is a viable pathway. However, although they both deploy GIN for graph modeling, in comparison to MoMu's adaptation with linear layers, MolCA's graph adaptation utilizing the Q-Former \cite{blip2} module achieves a much more significant improvement.

\begin{figure}[!ht]
\vskip -0.05in
\centering
\includegraphics[width=0.95\linewidth]{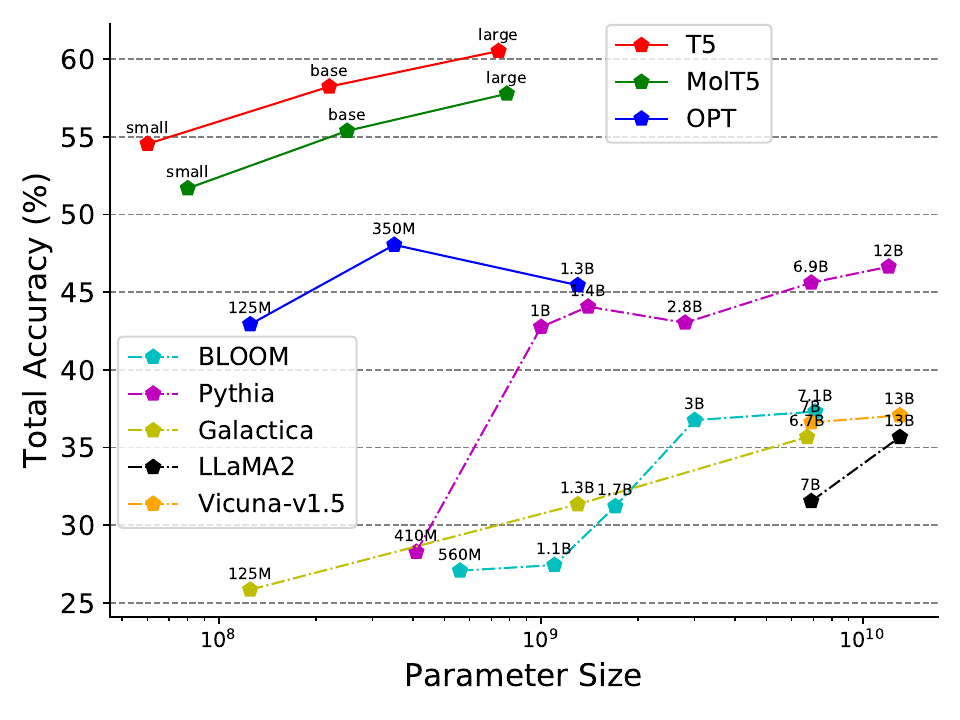}
\vskip -0.2in
\caption{\footnotesize{Model parameter size vs. total accuracy on \BenchName{} test set. Solid lines denote full fine-tune models, and dashed lines represent LoRA fine-tune.}}
\vskip -0.1in
\label{fig:scaling}
\vskip -0.1in
\end{figure}
%TODO: paste new figure

\subsection{Scaling Law for Molecular LLMs (RQ3)}
In \cref{fig:scaling}, we depict the performance variations of several models over increasing model scale. We observe a pronounced scaling effect across different training methods and model architectures, with the scale effect being more evident in the full fine-tuning approaches. This observation is consistent with previous analysis about the scale of parameters and indicates that scaling up model size is a promising way to enhance molecular modeling.

\vskip -0.1in
\section{Conclusion and Future Work}
\label{sec:Conclusion}
% Conclusion

In conclusion, this paper addresses the lack of evaluation for factual discrepancies in Large Language Models (LLMs) within the molecular domain. By organizing molecular descriptions into a taxonomy and constructing QA pairs through human and LLM efforts, we introduce \BenchName{}, a novel dataset for molecular factual question answering. Our evaluation reveals shortcomings in existing models, emphasizing critical factors for molecular comprehension and providing guidance for future LLM development.
Looking forward, we propose three key future directions:
(1) Design a powerful molecular model based on our analysis.
(2) Apply \BenchName{} in the training of molecular LLMs.
(3) Incorporate additional data sources to enrich \BenchName{}'s comprehensiveness.
% In this paper, we initially highlight the lack of evaluation for LLMs' factual discrepancies in the current molecular domain. 
% We then organize molecular descriptions into the taxonomy and construct QA pairs with human and LLM efforts. Finally, we present \BenchName{}, a novel dataset aimed at molecular factual question answering. Our evaluation uncovers several aspects where existing models exhibit sub-optimally and highlights critical factors for molecular comprehension, guiding the development of molecular LLMs.

% We propose below future directions: (1) Design a powerful molecular model following our analysis. (2) Apply \BenchName{} in the training of molecular LLMs. (3) Incorporate additional data sources to enrich \BenchName{}'s comprehensiveness.
% Future Work

\clearpage
\section*{Limitations}
\label{sec:Limitations}
% 1. data replenish: include more corpora for a more balanced dataset.
% 2. model trail: Try to combine several conclusions to build a more powerful method.
% 3. 
We conclude our limitations into the following aspects: (1) Imbalanced data distribution across different aspects, notably with Structure and Source data dominating the majority. This skew results from the overall prevalence of structural and source-related information in the data sources. To address this, future efforts will focus on introducing more data related to properties and applications while expanding topic coverage and diversity, all while safeguarding against data leakage. (2) Absence of full fine-tuning for large models: Under the constraint of computational resources, we fail to fully fine-tune LLMs with 7B parameters and above, leading us to opt for adaptation-based fine-tuning methods. (3) Absence of a specially designed molecular model: As highlighted in the Future Work, we did not provide a self-designed model based on benchmark analysis. In future endeavors, high-quality domain datasets and appropriate multi-modal fusion strategy will be leveraged to develop a molecular model with robust molecular comprehension.

%%% %%%
\section*{Potential Risks}
\label{sec:Ethical}
Although \BenchName{} offers a viable approach for factual assessment in the molecular domain with reliable data quality, there remains a risk of misuse. Evaluations on this dataset may not accurately represent a model's comprehension over all molecules. \BenchName{} could potentially be leveraged to furnish a veneer of reliability for models with underlying risks.

\bibliography{anthology,custom}
\bibliographystyle{acl_natbib}

\appendix
\section{Appendix}
\label{sec:appendix}

% License
\subsection{Data Sources and License}
As depicted in Table~\ref{tab:license}, we elaborate on the origins and legal permissions associated with each data component utilized in the development of the \BenchName{}. This encompasses both biomolecular data and textual descriptions. Thorough scrutiny was conducted on all data origins to confirm compatibility with our research objectives and subsequent utilization. Proper and accurate citation of these data sources is consistently maintained throughout the paper.

\renewcommand\arraystretch{1.3}
\begin{table*}[!ht]
    \centering
    \small
    % \vskip -0.2in
    % \vskip 0.07in
    \scalebox{0.97}{
    % \begin{tabular}{p{3.5cm}p{9cm}}
    \begin{tabular}{p{0.3cm}p{4cm}p{9.5cm}}
\toprule
\textsc{\textbf{Aspect}}  & \textsc{\textbf{Sub Topics}} & \textsc{\textbf{Leaf Topics}}    \\
\hline
\multicolumn{1}{l|}{\multirow{7}{*}{\textit{Property}}} & \textit{Biological and Pharmacological Activities}& "antimicrobial activity",
            "anti-neoplastic activity",
            "antioxidant activity",
            "enzyme inhibition",
            "ion channel activity",
            "receptor activity"... \\ \cline{2-3}
\multicolumn{1}{l|}{}  & \textit{Reaction Types} &                  "acetylation",
            "condensation",
            "dehydrogenation",
            "epoxidation",
            "glycosylation",
            "hydroxylation",
            "oxidation",
            "phosphorylation",
            "reduction"... \\ \cline{2-3}
\multicolumn{1}{l|}{}  & \textit{Chemical Interaction and Mechanism} & "action",
            "affinity",
            "binding",
            "conversion",
            "decomposition",
            "duration",
            "formation",
            "mechanism",
            "reaction/binding",
            "receptor affinity",
            "selectivity"... \\ \cline{2-3}
\multicolumn{1}{l|}{}  & \textit{Chemical Properties}   &        "chemical nature",
            "sensitivity",
            "ph value",
            "stability",
            "valence",
            "reactivity"  \\ \cline{2-3}
\multicolumn{1}{l|}{}  & \textit{Environmental and Safety Concerns}         &  "bio-accumulation",
            "xenobiotic",
            "cell permeability",
            "teratogenic agent",
            "environmental contaminant",
            "resistance",
            "safety concerns"  \\ \cline{2-3}
\multicolumn{1}{l|}{}  & \textit{Medical and Therapeutic Efficacy}          &  "analgesic activity",
            "anti-inflammatory activity",
            "antimalarial activity",
            "anti-mycobacterial",
            "carcinogenicity",
            "medical effects",
            "potency"...  \\ \cline{2-3}
\multicolumn{1}{l|}{}  & \textit{Physical and Sensory Properties}  & "abundance",
            "atomic mass",
            "boiling point",
            "color",
            "half-life",
            "odor",
            "optical activity",
            "physical state",
            "solubility",
            "taste",
            "volatileness"...   \\
\hline
\multicolumn{1}{l|}{\multirow{7}{*}{\textit{Application}}} & \textit{Agricultural Chemicals} & "fungicide",
            "herbicide",
            "insecticide",
            "disease control",
            "herbicide safener",
            "synthetic auxin",
            "phytoestrogen"... \\ \cline{2-3}
\multicolumn{1}{l|}{}  &  \textit{Biological Agents} & "antibiotic",
            "antifungal drug",
            "antibacterial drug",
            "antiprotozoal",
            "antiviral drug",
            "nematicide",
            "acaricide",
            "antiseptic"... \\ \cline{2-3}
\multicolumn{1}{l|}{}  & \textit{Chemical Applications and Techniques}   & "reagent",
            "indicator",
            "detection",
            "derivatisation agent",
            "fluorescent dye",
            "production",
            "chromatographic reagent",
            "tracer",
            "solvent",
            "food additive"...  \\ \cline{2-3}
\multicolumn{1}{l|}{}  & \textit{Pharmacodynamics and Pharmacokinetics}   &  "inhibitor",
            "antagonist",
            "prodrug",
            "modulator",
            "sympathomimetic agent",
            "allergen",
            "sodium channel blocker",
            "ligand",
            "agonist"... \\ \cline{2-3}
\multicolumn{1}{l|}{}  & \textit{Regulatory Status and Approval}   &  "approval",
            "withdrawn from market",
            "registered in"... \\ \cline{2-3}
\multicolumn{1}{l|}{}  & \textit{Research and Development}  & "experimental",
            "biomarker",
            "clinical development",
            "testing"...  \\ \cline{2-3}
\multicolumn{1}{l|}{} & \textit{Therapeutic Use} & 
            "anti-arrhythmia drug",
            "anti-allergic agent",
            "anti-asthmatic drug",
            "anticoronaviral agent",
            "anti-neoplastic agent",
            "anti-ulcer drug",
            "anti-HIV agent",
            "orphan drug",
            "recreational drug",
            "vasodilator"... \\ \hline
\multicolumn{1}{l|}{\multirow{4}{*}{\textit{Source}}} & \textit{found in} & "found in" \\ \cline{2-3}
\multicolumn{1}{l|}{} & \textit{metabolite} & "metabolite" \\ \cline{2-3}
\multicolumn{1}{l|}{} & \textit{derives from} & "derives" \\ \cline{2-3}
\multicolumn{1}{l|}{} & \textit{isolated from} & "isolated" \\ 
\hline
\multicolumn{1}{l|}{\multirow{6}{*}{\textit{Structure}}} & \textit{Biochemical and Biological Terms} & "active metabolite",
            "alkaloid"
            "coenzyme a",
            "enzyme",
            "epitope",
            "fatty acyl coa",
            "glucoside",
            "hapten",
            "nucleobase",
            "oligosaccharide",
            "sphingoid base",
            "substrate"... \\ \cline{2-3}
\multicolumn{1}{l|}{} & \textit{Chemical Bonding and Interactions} &  "glycosidic bond",
            "disulfide bonds",
            "double bond",
            "exocyclic double bond",
            "peptide bond",
            "c=c double bond",
            "bond",
            "connection",
            "attachment"... \\  \cline{2-3}
\multicolumn{1}{l|}{} & \textit{Chemical Compounds and Classes} & "acid",
            "alcohol",
            "amine",
            "cation",
            "dimer",
            "enamide",
            "hydrochloride",
            "ion",
            "lactam",
            "polyphenol",
            "salt",
            "phosphate",
            "sulfate",
            "oxoanion",
            "zwitterion"... \\ \cline{2-3}
\multicolumn{1}{l|}{} & \textit{Chemical Species and States} & "anhydrous form",
            "heptahydrate form",
            "oxidation state",
            "hydrate",
            "major microspecies",
            "deoxygenated",
            "major species",
            "microspecies"... \\ \cline{2-3}
\multicolumn{1}{l|}{} & \textit{Functional Groups and Chemical Entities} & 
            "acyl group",
            "alcohol group",
            "alkyl group",
            "anilino group",
            "carbamoyl group",
            "chloro group",
            "epoxy group",
            "ester group",
            "fatty acyl group",
            "hydrazino group",
            "hydroperoxy group",
            "isopropyl substituent",
            "keto group",
            "methyl group",
            "oxo group",
            "pentyl group",
            "phosphate group",
            "primary hydroxy group",
            "s-acyl component",
            "s-methyl group",
            "sulfo group",
            "thiol group"... \\ \cline{2-3}
\multicolumn{1}{l|}{} & \textit{Molecular Structure and Configuration} & 
            "alpha-branch",
            "alpha-carbon",
            "backbone",
            "branch",
            "bridge",
            "core",
            "composition",
            "configuration",
            "linked group",
            "n-substituent",
            "oh groups",
            "omega-hydroxy",
            "position",
            "prenyl units",
            "terminal",
            "terminal group",
            "glycosyl fragment",
            "repeating unit",
            "sequence",
            "subcomponents",
            "side chain",
            "nucleus",
            "sugar fragment",
            "unit"... \\
\bottomrule
    \end{tabular}
    }
    \caption{
    \footnotesize
    Taxonomy of \textit{Property}, \textit{Application}, \textit{Structure} and \textit{Source} aspects in \BenchName{}. \textbf{Leaf Topics} correspond to the most granular concepts, while \textbf{Sub Topics} aggregate leaf topics further. The table presents only a subset of leaf topics.}
    \vskip 0.3in
    \label{tab:taxonomy_property_app_source_structure}
\end{table*}

\begin{table*}[ht]
    \centering
    \vskip 0.1in
    \scalebox{0.69}{
    \begin{tabular}{p{4cm}p{7cm}p{11cm}}
    \toprule
    \textsc{\textbf{Data Sources}} & \textsc{\textbf{License URL}} & \textsc{\textbf{License Note}} \\
    \midrule
    PubChem & \url{https://www.nlm.nih.gov/web_policies.html} & Works produced by the U.S. government are not subject to copyright protection in the United States. Any such works found on National Library of Medicine (NLM) Web sites may be freely used or reproduced without permission in the U.S. \\
    FDA Pharm Classes & \url{https://www.fda.gov/about-fda/about-website/website-policies} & Unless otherwise noted, the contents of the FDA website (www.fda.gov), both text and graphics, are not copyrighted. They are in the public domain and may be republished, reprinted and otherwise used freely by anyone without the need to obtain permission from FDA. Credit to the U.S. Food and Drug Administration as the source is appreciated but not required. \\
    Drug Bank & \url{https://creativecommons.org/licenses/by-nc/4.0/legalcode} & Subject to the terms and conditions of this Public License, the Licensor hereby grants You a worldwide, royalty-free, non-sublicensable, non-exclusive, irrevocable license to exercise the Licensed Rights in the Licensed Material to: reproduce and Share the Licensed Material, in whole or in part, for NonCommercial purposes only; and produce, reproduce, and Share Adapted Material for NonCommercial purposes only. \\
    ChEBI & \url{https://creativecommons.org/licenses/by/4.0/} & You are free to: Share — copy and redistribute the material in any medium or format. Adapt — remix, transform, and build upon the material for any purpose, even commercially. \\
    LOTUS  & \url{https://lotus.nprod.net/} & LOTUS is one of the biggest and best-annotated resources for natural products occurrences available free of charge and without any restriction. \\
    CAMEO Chemicals & \url{https://cameochemicals.noaa.gov/help/reference/terms_and_conditions.htm?d_f=false} & CAMEO Chemicals and all other CAMEO products are available at no charge to those organizations and individuals (recipients) responsible for the safe handling of chemicals. \\
    Toxin-Toxin-Target Database (T3DB) & 
    \url{http://www.t3db.ca/} & T3DB is offered to the public as a freely available resource. Use and re-distribution of the data, in whole or in part, for commercial purposes requires explicit permission of the authors and explicit acknowledgment of the source material (T3DB) and the original publication. \\  
    \bottomrule
    \end{tabular}
    
    }
    \caption{
    \footnotesize
    Data resources and licenses utilized in data collection for \BenchName{}.
    }
    \vskip -0.1in
    \label{tab:license}
\end{table*}

\subsection{Details about Taxonomy}
\begin{figure*}[!htp]
    \centering
    \includegraphics[width=0.98\linewidth]{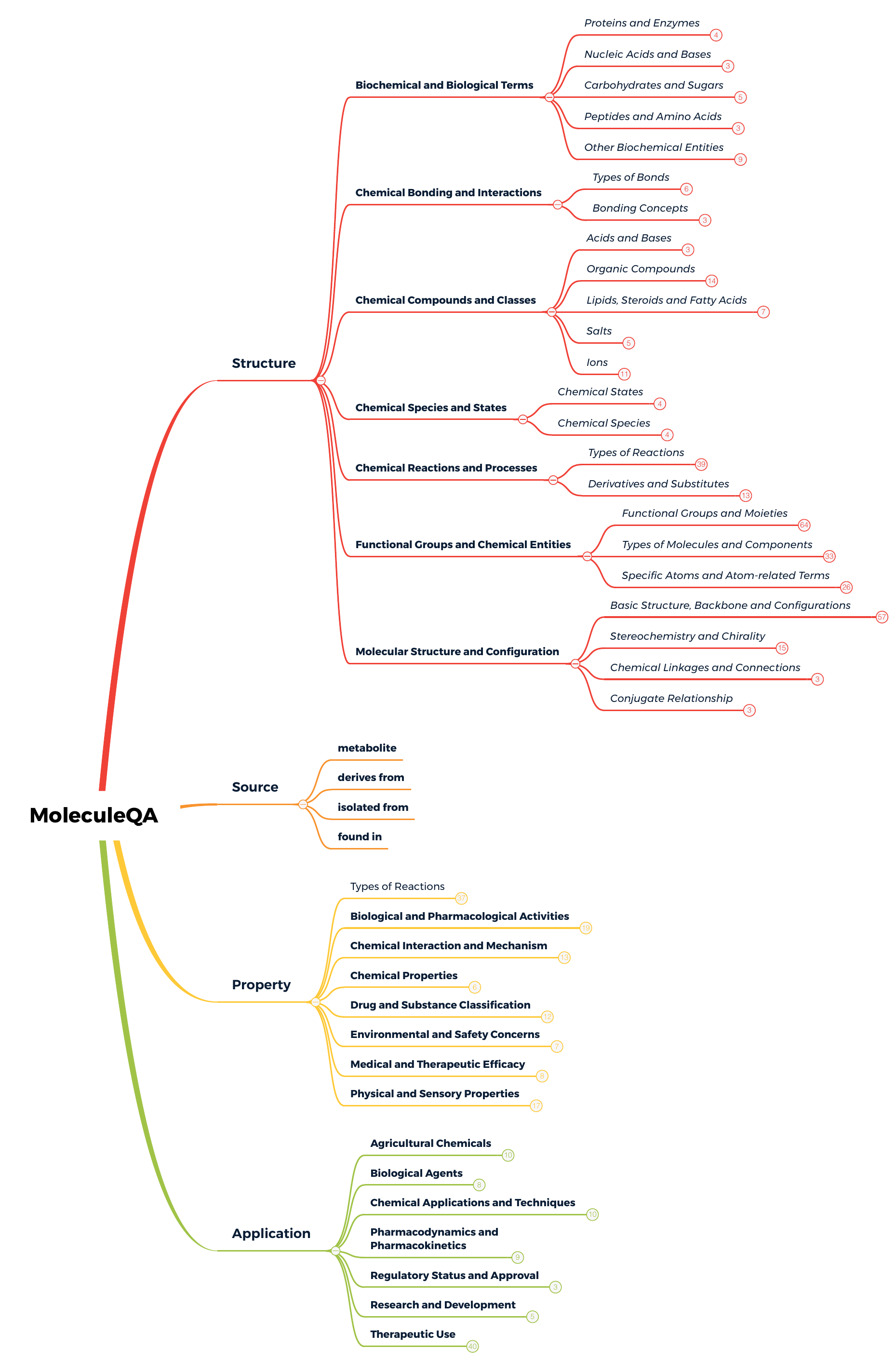}
    \vskip -0.1in
    \caption{\footnotesize{
The overarching structure of the \BenchName{} taxonomy comprises multiple aspects and subtopics arranged hierarchically to categorize various facets of molecular factual knowledge. Due to space constraints, we did not elaborate on all leaf topics.}}
    \label{fig:taxomomy_all}
    \vskip -0.1in
\end{figure*}

We present the overall hierarchical structure of the taxonomy upon which \BenchName{} is based in Figure \ref{fig:taxomomy_all}. Additionally, Table \ref{tab:taxonomy_property_app_source_structure} provides details regarding the subtopics and part of leaf topics encompassed within each of the four aspects: \textit{Structure}, \textit{Source}, \textit{Property}, and \textit{Application}.

% layout

\subsection{Experimental Setup Details}
\subsubsection{Baselines}
The following parts will individually introduce the models we evaluated in this study and the approaches used for implementation.

\noindent
\textbf{T5} \cite{T5} is an encoder-decoder model pre-trained on a multi-task mixture of unsupervised and supervised tasks for which each task is converted into a text-to-text format. We directly fine-tuned it on \BenchName{} dataset from public checkpoints \footnote{\url{https://github.com/google-research/text-to-text-transfer-transformer/blob/main/released_checkpoints.md\#t511}} with three different model sizes: small, base and large. It's important to note that the original T5 pre-training does not incorporate any specific knowledge related to the domain of molecules.

\noindent
\textbf{MolT5} \cite{MolT5} undergoes joint training on molecule SMILES from the ZINC-15 dataset \cite{ZINC15} and a general corpus from the C4 dataset \cite{T5}, enabling MolT5 to acquire prior knowledge in both of these domains. It contains three different sizes: small, base, and large. 
In the experiment, we utilized pre-trained model checkpoints of various sizes~\footnote{\url{https://huggingface.co/laituan245/molt5-small}, \url{https://huggingface.co/laituan245/molt5-base/}, \url{https://huggingface.co/laituan245/molt5-large/}} released by the authors. Subsequently, we conducted full fine-tuning on the \BenchName{} train set, followed by evaluating on the test set.

\noindent
\textbf{MoMu} \cite{MoMu} is pre-trained using molecular 2D graphs and their semantically related textual data (crawled from published Scientific Citation Index papers) via contrastive learning. 
We adopted MoMu-K pre-trained checkpoints \footnote{\url{https://github.com/ddz16/MoMu?tab=readme-ov-file\#pretrain}} where the text encoder is initialized with the weights of KV-PLM \cite{KV-PLM}. Following the original methodology, we injected encoded graph features into MolT5-base \& large and conducted finetuning on \BenchName{}.

\noindent
\textbf{BioT5} \cite{BioT5} as a comprehensive pre-training framework, builds upon the methodology of MolT5 while enhancing cross-modal integration into biology through chemical knowledge and natural language associations. It leverages SELFIES for robust molecular representations and extracts knowledge from the surrounding context of bio-entities in unstructured biological literature. We utilized the official base version pre-trained checkpoint \footnote{\url{https://huggingface.co/QizhiPei/biot5-base}} and converted the \BenchName{} data into the corresponding format for fine-tuning.

\noindent
\textbf{MolCA} \cite{MolCA} facilitates a language model (LM), such as Galactica, in comprehending both text- and graph-based molecular contents through its cross-modal projector. This projector, implemented as a Q-Former, serves to bridge the representation space of a graph encoder with the text space of an LM. Additionally, MolCA employs a uni-modal adapter to enable efficient adaptation of the LM to downstream tasks. 
We conducted pre-training, including both stage 1 and stage 2, on the 125M and 1.3B versions, based on the official code and cleaned data \footnote{\url{https://github.com/acharkq/MolCA}}. Subsequently, we performed finetuning on \BenchName{}.

\noindent
\textbf{BioMedGPT-LM-7B} \cite{BioMedGPT} It is a large generative language model based on Llama2 in the biomedical domain. It was fully fine-tuned from the Llama2-7B-Chat with millions of biomedical papers from the S2ORC corpus \cite{S2ORC}. We directly apply the LoRA finetuning method on the checkpoint \footnote{\url{https://huggingface.co/PharMolix/BioMedGPT-LM-7B}} provided by the official source.

\noindent
\textbf{OPT} \cite{OPT} is a series of open-sourced large causal language models which perform similar in performance to GPT-3 \cite{GPT-3}. For comparison with fully fine-tuned T5 series models, we opted to fully fine-tune OPT-125M, -350M, and -1.3B size models on \BenchName{}. In our implementation, we referred to the interfaces provided by Hugging Face \footnote{\url{https://huggingface.co/docs/transformers/model_doc/opt}}.

\noindent
\textbf{GALACTICA} \cite{Galactica} is a large language model (LLM) for Science: trained on over 48 million papers, textbooks, reference material, compounds, proteins and other sources of scientific knowledge. 
We selected GALACTICA-125M, -1.3B, and -7.1B versions of the model \footnote{\url{https://huggingface.co/models?other=galactica}} and conducted fine-tuning using LoRA on \BenchName{}.

\noindent
\textbf{Pythia} \cite{Pythia} is an open suite of large language models, all trained on public data in the same order. These models vary in size, ranging from 70M to 12B parameters. They were trained on the Pile dataset, which is constructed from 22 diverse high-quality subsets. 
We opted to conduct finetuning based on LoRA on the standard versions of Pythia-410M, -1B, -2.8B, -6.9B, and -12B sizes models ~\footnote{\url{https://huggingface.co/models?other=pythia}}.

\noindent
\textbf{BLOOM} \cite{BLOOM} is an autoregressive large language model, trained to continue text from a prompt on vast amounts of text data using industrial-scale computational resources. It was trained on the ROOTS \cite{ROOTS} corpus, a dataset comprising hundreds of sources in 46 natural and 13 programming languages (59 in total). For model scaling evaluation, we chose to conduct finetuning based on LoRA on the BLOOM-560M, -1.1B, -1.7B, -3B, and -7.1B sizes versions of the model ~\footnote{\url{https://huggingface.co/docs/transformers/model_doc/bloom}}. Subsequently, we provided the results on the \BenchName{} test set.

\noindent
\textbf{LLaMA-2} \cite{Llama2} is a collection of large language models with parameters ranging from 7 billion to 70 billion. The model architecture remains largely unchanged from that of LLaMA-1 models \cite{Llama}, but 40\% more data was used to train the foundational models. Specifically, Llama 2 includes pre-trained and fine-tuned models optimized for dialogue applications, termed Llama 2-Chat. 
We opted to utilize the LLaMA-2-Chat 7B and 13B models \footnote{\url{https://huggingface.co/docs/transformers/model_doc/llama2}} and transformed \BenchName{} into instruction samples for LoRA fine-tuning.

\noindent
\textbf{Vicuna-v-1.5} \cite{vicuna} is an open-source chatbot that has been trained by fine-tuning LLaMA on over 150K user-shared conversations collected from ShareGPT.com. Preliminary evaluation, conducted with GPT-4 as the judge, demonstrates that the Vicuna series achieves competitive performance when compared to OpenAI ChatGPT, while also outperforming other models such as LLaMA. We selected the v1.5 series models and conducted LoRA Finetuning on both the 7B and 13B versions \footnote{\url{https://huggingface.co/lmsys/vicuna-7b-v1.5}, \url{https://huggingface.co/lmsys/vicuna-13b-v1.5}}.

\noindent
\textbf{Mol-Instructions-7B} \cite{Mol-Instructions} is a low-rank adapter designed for LLaMA-2 base LLM, specifically trained on molecule-oriented instructions sourced from the Mol-Instructions dataset. We utilize the version tailored for LLaMA-2-Chat \footnote{\url{https://huggingface.co/zjunlp/llama2-molinst-molecule-7b}}, merging the adapter back to the base LLM before proceeding with LoRA fine-tuning.

\noindent
\textbf{Mixtral-8×7B} \cite{Mixtral8x7B} is a Sparse Mixture of Experts (SMoE) language model consisting of a decoder-only architecture. Its feedforward block selects from a set of 8 distinct groups of parameters. Notably, it is recognized as the most robust open-weight model currently available, licensed under Apache 2.0. We adopt a locally deployed approach for conducting few-shot prompting inference.

\noindent
\textbf{GPT-3.5-turbo and GPT-4.} 
For closed-source models such as OpenAI GPT Family GPT-3.5-turbo \cite{Chatgpt} and GPT-4 \cite{gpt4v}, we employ batch inference via APIs for conducting few-shot prompt inference. This approach significantly enhances evaluation efficiency and reduces overhead.

\subsubsection{Hyper-parameters}
For MolT5, MoMu, T5, and BioT5, we employed the original codebases and hyper-parameters provided in the respective papers for full fine-tuning. Specifically, these models were trained on a single NVIDIA 48GB A6000 GPU. Except for BioT5, which had a learning rate set to 1e-3, the learning rates for all other models were set to 1e-4. All models underwent fine-tuning for 100 epochs on the training set, and the checkpoint with the best performance on the development set was selected for evaluation on the test set. 

For MolCA, we utilized the author's recently updated dataset (excluding any data leakage concerns) and conducted pre-training stage 1 and stage 2 training on 2 NVIDIA 48GB A6000 GPUs. We maintained consistency with the training hyper-parameters provided in the original paper. Subsequently, we fine-tuned pre-trained checkpoints of different sizes on \BenchName{}, with a total batch size set to 16. The 125M model was trained on a single GPU card, while the 1.3B model was trained on two GPU cards. The fine-tuning total epochs were set to 100 for all versions.

For full fine-tuning of the OPT series, we conducted training on 4 A6000 GPUs for the 125M and 350M versions and 8 GPUs for the 1.3B version. The total batch size was set to 256, and the learning rates were set to 3e-4 and 2e-4 for the respective versions. All other hyper-parameters were kept consistent with those specified in the original paper. We performed full fine-tuning for 60 epochs, as we observed over-fitting phenomena when exceeding 50 epochs.

For the remaining experiments based on LoRA tuning, we employed the Alpaca-LoRA codebase for instruction fine-tuning. Except for the 13B size model trained on 8 A6000 GPUs, all other models were trained on 4 GPUs. The total batch size was set to 400, with gradient accumulation and learning rate adjusted according to the model size (typically set to 3e-4). We set the total training epochs to 20. 

Regarding the LoRA configuration, we utilized the PEFT \footnote{\url{https://github.com/huggingface/peft}} library for implementation. We set LoRA's rank $r$ as 16, $\alpha$ as 16, dropout rate as 0.05, and applied LoRA to all modules of [\texttt{"q/k/v/o\_proj"}, \texttt{"gate\_proj"}, \texttt{"down/up\_proj"}] (adjusting module names if necessary based on actual implementation). Equivalent trainable parameters are reported in Table \ref{tab:main}.

\subsection{Scaling Law in Detail}
% We illustrate the accuracy rate changes across different series models as the model parameter scale increases, focusing on the four main aspects: \textit{Structure} (Table \ref{fig:scaling_structure}), \textit{Source} (Table \ref{fig:scaling_source}), \textit{Property} (Table \ref{fig:scaling_property}), and \textit{Application}(Table \ref{fig:scaling_application}). 

\begin{figure*}[!htbp]
    \centering
    \includegraphics[width=1.0\linewidth]{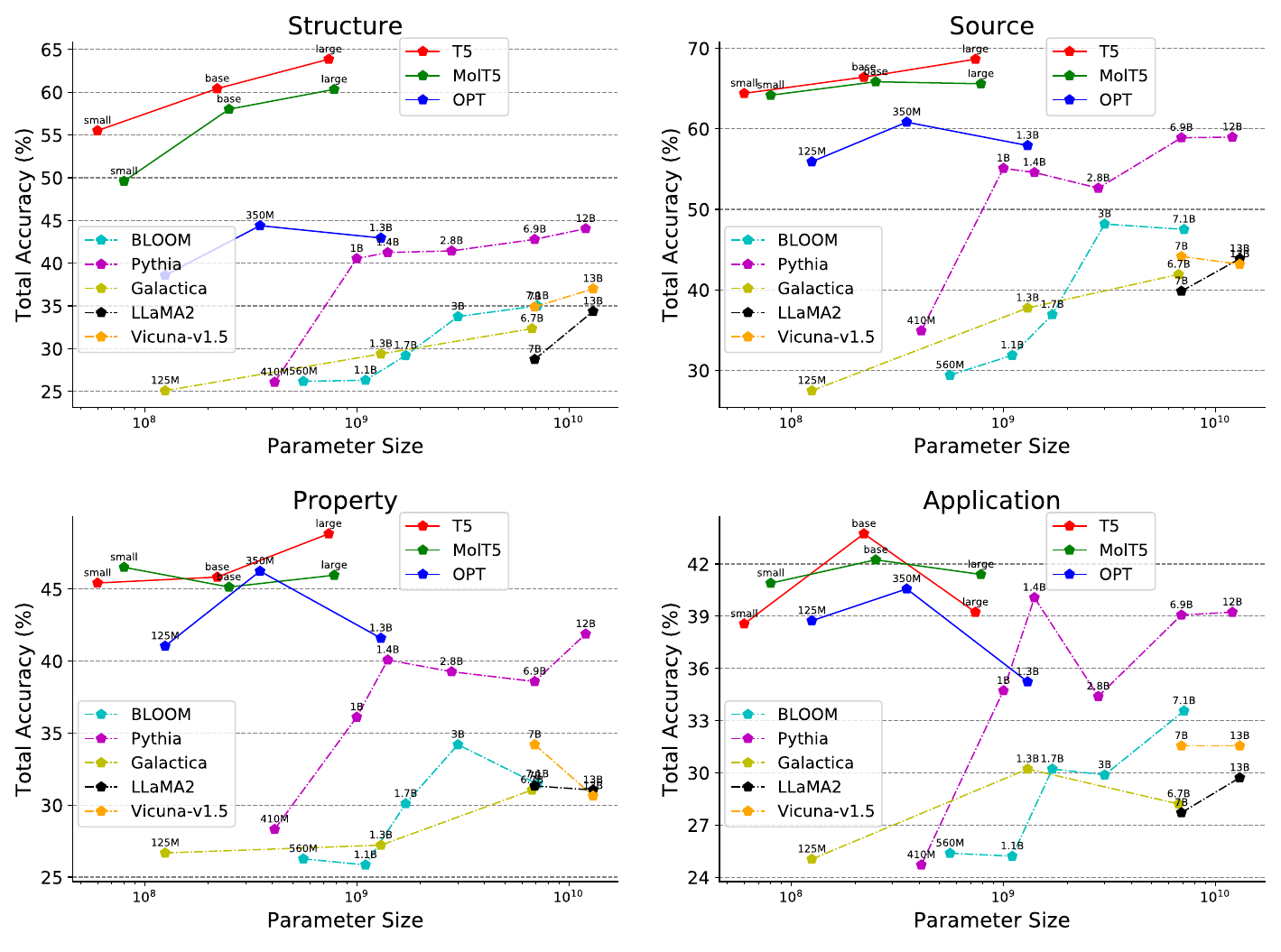}
    \vskip -0.1in
    \caption{\footnotesize{Model parameter size vs. Accuracy in four aspects.}}
    \label{fig:scaling_detail}
    \vskip -0.5in
\end{figure*}

In Figure \ref{fig:scaling_detail}, we illustrate the accuracy rate changes across different series models as the model parameter scale increases, focusing on the four main aspects: \textit{Structure}, \textit{Source}, \textit{Property}, and \textit{Application}.

For fully fine-tuned models, we conducted comparisons between T5-based models (represented by T5 and MolT5) and decoder-only-based models (represented by OPT). To validate whether adaptively fine-tuning can augment general domain LLMs to acquire molecular domain factual knowledge, we compared models such as BLOOM, Pythia, GALACTICA, and the LLaMA2-series models using LoRA fine-tuning.

% \begin{figure}[!htbp]
%     \centering
%     \includegraphics[width=1.05\linewidth]{fig/model_scaling/structure.pdf}
%     \vskip -0.1in
%     \caption{\footnotesize Model parameter size vs. accuracy in \textit{Structure}.}
%     \label{fig:scaling_structure}
%     \vskip -0.1in
% \end{figure}

% \begin{figure}[!htbp]
%     \centering
%     \includegraphics[width=1.05\linewidth]{fig/model_scaling/source.pdf}
%     \vskip -0.1in
%     \caption{\footnotesize Model parameter size vs. accuracy in \textit{Source}.}
%     \label{fig:scaling_source}
%     \vskip -0.1in
% \end{figure}

% \begin{figure}[!htbp]
%     \centering
%     \includegraphics[width=1.05\linewidth]{fig/model_scaling/property.pdf}
%     \vskip -0.1in
%     \caption{\footnotesize Model parameter size vs. accuracy in \textit{Property}.}
%     \label{fig:scaling_property}
%     \vskip -0.1in
% \end{figure}

% \begin{figure}[!htbp]
%     \centering
%     \includegraphics[width=1.05\linewidth]{fig/model_scaling/application.pdf}
%     \vskip -0.1in
%     \caption{\footnotesize Model parameter size vs. accuracy in \textit{Application}.}
%     \label{fig:scaling_application}
%     \vskip -0.1in
% \end{figure}

\subsection{Prompt of Different Tasks for LLM}
In the construction process of \BenchName{}, we deploy LLMs to finish the following tasks: (1) Corpus classification;  (2) Topic extraction; (3) Answer generation; (4) Semantic consistency validation. We report the definitions and task contexts,  which are components of prompts for LLMs, of these tasks in \cref{tab:appendix-task-define}.
% Task Definition

\renewcommand\arraystretch{1.5}
\begin{table*}[ht]
\centering
\footnotesize
\scalebox{1.0}{
\begin{tabular}{p{2.8cm}p{4.5cm}p{7cm}}
\toprule
\textbf{\textsc{Task}} & \textbf{\textsc{Definition}} & \textbf{\textsc{Task Context}} \\ \hline
\textit{Corpus Classification}  & \makecell[l]{Classify molecular descriptions \\from the data source into one of four \\aspects.} & \makecell[l]{You are a research assistant for molecular research.\\
Please help me to classify some corpus. \\
Four kinds of content are included in this corpus :\\ The first is Source, which describes... \\
The second is... } \\
\hline
\textit{Topic Extraction}  & \makecell[l]{Extract attributes of molecules in \\specific aspect from original \\descriptions.} & \makecell[l]{You are a chemical research assistant, \\you are familiar with description text of molecules,
\\you need to help me extract molecules' Source\\information, which describes...} \\
\hline

\textit{Answer Generation}  & \makecell[l]{Generate answer for given \\question with original description} & \makecell[l]{
You are a chemistry research assistant, and I need you \\to complete the following task: You will be given a \\detailed description of a molecule and a question, please \\extract specific information from the given description \\to answer the question... } \\
\hline

\textit{Semantic Consistency Validation} & \makecell[l]{Check if generated answer \\has consistent semantic \\with original description.} & \makecell[l]{
You are a chemistry research assistant, and I need you \\to complete the following task: You will be given a \\description of a molecule and a sentence transcribed from\\it, please justify whether their semantics are consistent... } \\
\bottomrule
\end{tabular}
}
\caption{
    Definition and context for each task. We prompt LLMs to finish these tasks for \BenchName{} construction.
}
\label{tab:appendix-task-define}
\end{table*}

\subsection{Few-Shot Details and Prompt Exhibition}
We introduce details about our few-shot setting: For each aspect, we select a representative and various samples from different topics as examples to construct an aspect-specific 10-shot prompt. We demonstrate selected samples and the format of prompt in \textit{Source} aspect in \cref{tab:appendix-GPT-Prompt}.

% LLM Prompts
\begin{table*}[h!]\centering 
\footnotesize
\begin{minipage}{\textwidth}\vspace{0mm}    
\centering
\begin{tcolorbox} 
    \centering
    \small
     % \hspace{-6mm}
\begin{tabular}{p{1.0\textwidth}}
\VarSty{messages} = [
            \{\var{"role":"system", "content":} \var{f"""}\\You are a chemistry research assistant, and I'd like to test your professional ability on molecule understanding, please complete the following task:\\
            You are provided with the SMILES representation of a molecule and asked a question about the molecule's \texttt{source}-related knowledge (\texttt{Source} means the natural or synthetic origin, as well as
the production context related to a molecule), with four options given.  Three of these options do not describe the given molecule, and you must select the correct option. 
            \\ 
% \textbf{The background of the dataset and task is shown below:} \\
% The Blood-brain barrier penetration (BBBP) dataset comes from a recent study on the modeling and prediction of barrier permeability. As a membrane separating circulating blood and brain extracellular fluid, the blood-brain barrier blocks most drugs, hormones, and neurotransmitters. Thus penetration of the barrier forms a long-standing issue in the development of drugs targeting the central nervous system.   \\
Here are several examples to show how to finish the Question Answering task: \\
\#\#\# \\
Example 1: \\
\texttt{Molecular SMILES}: C1=CC(=CC=C1/C=CC(=O)O[C@@H]([C@H](C(=O)O)O)C(=O)O)O \\
\texttt{Question}: Which molecule does this molecule derive from? \\ 
\texttt{Choices}: \\
A: It derives from a meso-tartaric acid and a cis-4-coumaric acid. \\
B: It derives from a meso-tartaric acid and a cis-caffeic acid. \\
C: It derives from a cyanidin cation and a cis-4-coumaric acid. \\
D: It derives from a cis-vaccenic acid and an oleic acid.
\\
\texttt{Answer}: A \\
\#\#\# 
\\ \\

\#\#\# \\
Example 2: \\
\texttt{Molecular SMILES}: COC1=C(C=C(C=C1)C=O)OC \\
\texttt{Question}: Where this molecule can be found? \\ 
\texttt{Choices}: \\
A: It can be found in leaves and fruit of cowberry Vaccinium vitis-idaea, grape seeds and beer. \\
B: It can be found in peppermint, ginger, raspberry, and other fruits. \\
C: It can be found in edible vegetables, grains, and fruits. \\
D: It can be found in grape seeds, in Hibiscus cannabinus (kenaf) root and bark, in apple and in cacao.
\\
\texttt{Answer}: B \\
\#\#\# 
\\...
\\\\

Notice that here are some rules you need to follow: \\
1. Your answer for each question should be one of A/B/C/D, which corresponds to the four options. \\
2. For my convenience, please give me a list of ANSWERs for the given instances in the format 'Answer X: ...', without any other information. \\

\var{"""}\}\\

 \{
 \var{"role":"user", "content":} \var{f"""}\\
Please give me your choices for these instances in the above examples' styles. No other information is required. \\
\texttt{Instance ID}: <Instance ID> \\
\texttt{Molecular SMILES}: <Instance SMILES> \\
\texttt{Question}: <Instance Question>\\
\texttt{Choices}: <Instance Choices>\\
\var{"""}\}\\
]
\\
\end{tabular}
\end{tcolorbox}
\vspace{-2mm}
\caption{An illustration depicting the process of constructing few-shot in-context-learning prompts for \BenchName{} test set inference with GPT-4-like large-scale universal models.}
\label{tab:appendix-GPT-Prompt}
\end{minipage}
\end{table*}

\end{document}